\documentclass[12pt]{article}
\usepackage{fullpage,archive}
\usepackage{times}

\usepackage[dvips]{graphics}

\usepackage{epsfig}
\usepackage{graphicx}
\usepackage{wrapfig}
\usepackage{amsfonts}
\usepackage{amsthm}
\usepackage{multirow}
\usepackage{amsmath}
\usepackage{rotating}
\usepackage{tabularx}
\usepackage{subfigure}

\usepackage{fancyhdr}
\fancypagestyle{rcsfooters}{%
\fancyhead[L]{Submitted to {\it The American Journal of Human Genetics}}

}

\title{A Multivariate Regression Approach to Association Analysis of Quantitative Trait Network}

\author{Seyoung Kim \and Kyung-Ah Sohn \and Eric P. Xing}

\date{}

\abstract{
Many complex disease syndromes such as asthma consist of a large
number of highly related, rather than independent, clinical
phenotypes, raising a new technical challenge in identifying genetic
variations associated simultaneously with correlated traits. In this
study, we propose a new statistical framework called graph-guided
fused lasso (GFlasso) to address this issue in a principled way. Our
approach explicitly represents the dependency structure among the
quantitative traits as a network, and leverages this trait network
to encode structured regularizations in a
multivariate regression model over the genotypes and traits, so that
the genetic markers that jointly influence subgroups of highly correlated
traits can be detected with high sensitivity and
specificity. While most of the traditional methods examined each
phenotype independently and combined the results afterwards, our
approach analyzes all of the traits jointly in a single statistical
method, and borrow information across correlated phenotypes to
discover the genetic markers that perturbe a subset of correlated
triats jointly rather than a single trait. Using simulated datasets
based on the HapMap consortium data and an asthma dataset, we compare the
performance of our method with the single-marker analysis, and other
sparse regression methods such as the ridge regression and the lasso
that do not use any structural information in the traits. Our
results show that there is a significant advantage in detecting the
true causal SNPs when we incorporate the correlation pattern in
traits using our proposed methods.
}


\trnumber{Technical Report\\ CMU-ML-08-113}

\begin{document}

\maketitle

\newcommand{\yv}{\mathbf{y}}
\newcommand{\Yv}{\mathbf{Y}}
\newcommand{\xv}{\mathbf{x}}
\newcommand{\Xv}{\mathbf{X}}
\newcommand{\D}{\scriptscriptstyle}

\newcommand{\bm}[1]{\mbox{\boldmath$#1$\unboldmath}}
\newcommand{\mysgn}{\textrm{sign}}

\newcommand{\Gwa}{{${\rm G_wFlasso}$}}
\newcommand{\Gw}{{${\rm G^1_wFlasso}$}}
\newcommand{\Gww}{{${\rm G^2_wFlasso}$}}
\newcommand{\Gc}{{${\rm G_cFlasso}$}}

\section{Introduction}

\begin{figure}[t!]\centering
\begin{picture}(350,350)
\setlength{\unitlength}{1cm}
\put(-1.5,12.3){
\includegraphics[scale = 0.87,angle=-90, bb=115 150 500 650]{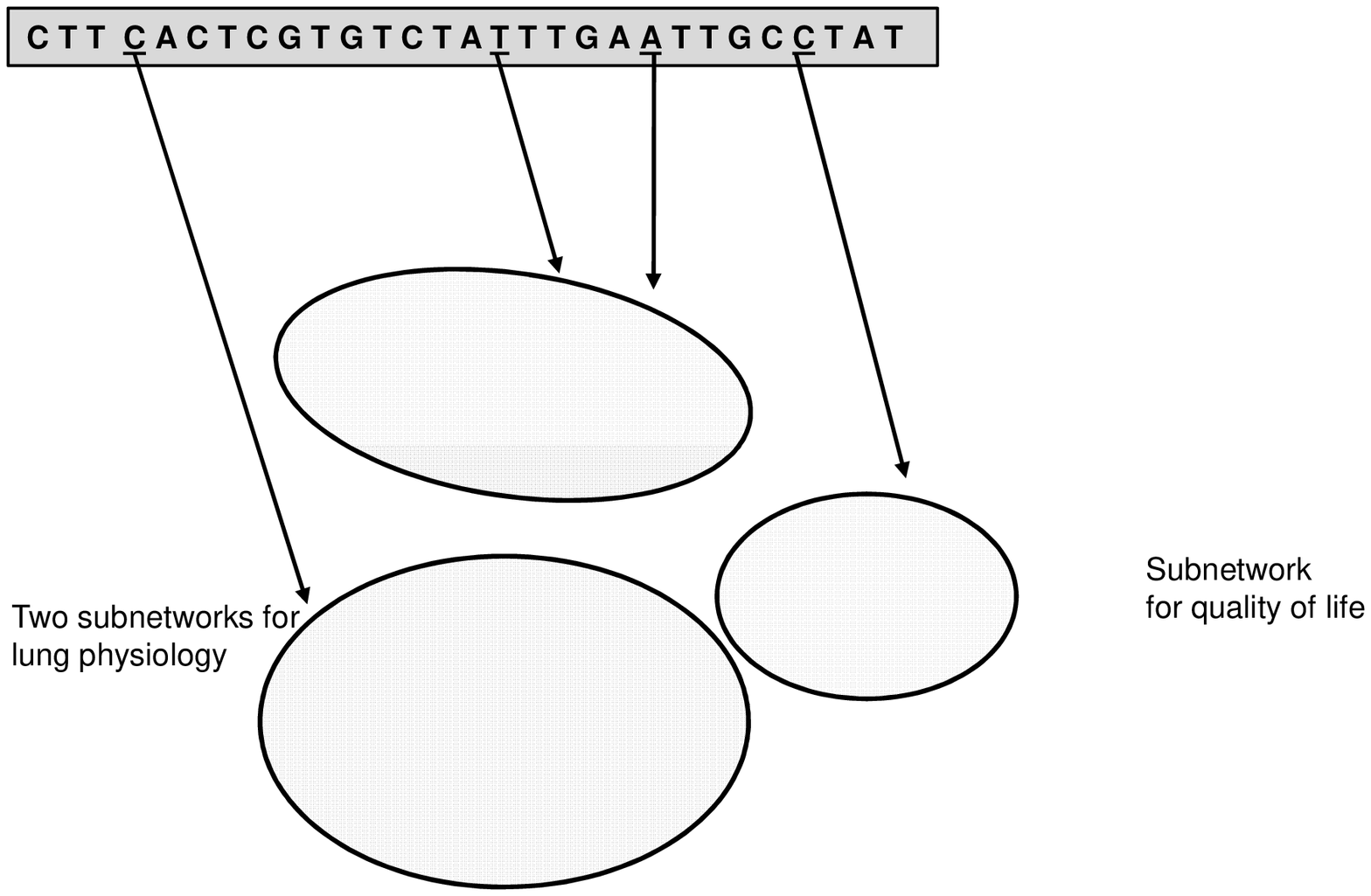}}
\put(1,0){\includegraphics[scale = 0.65,angle=0]{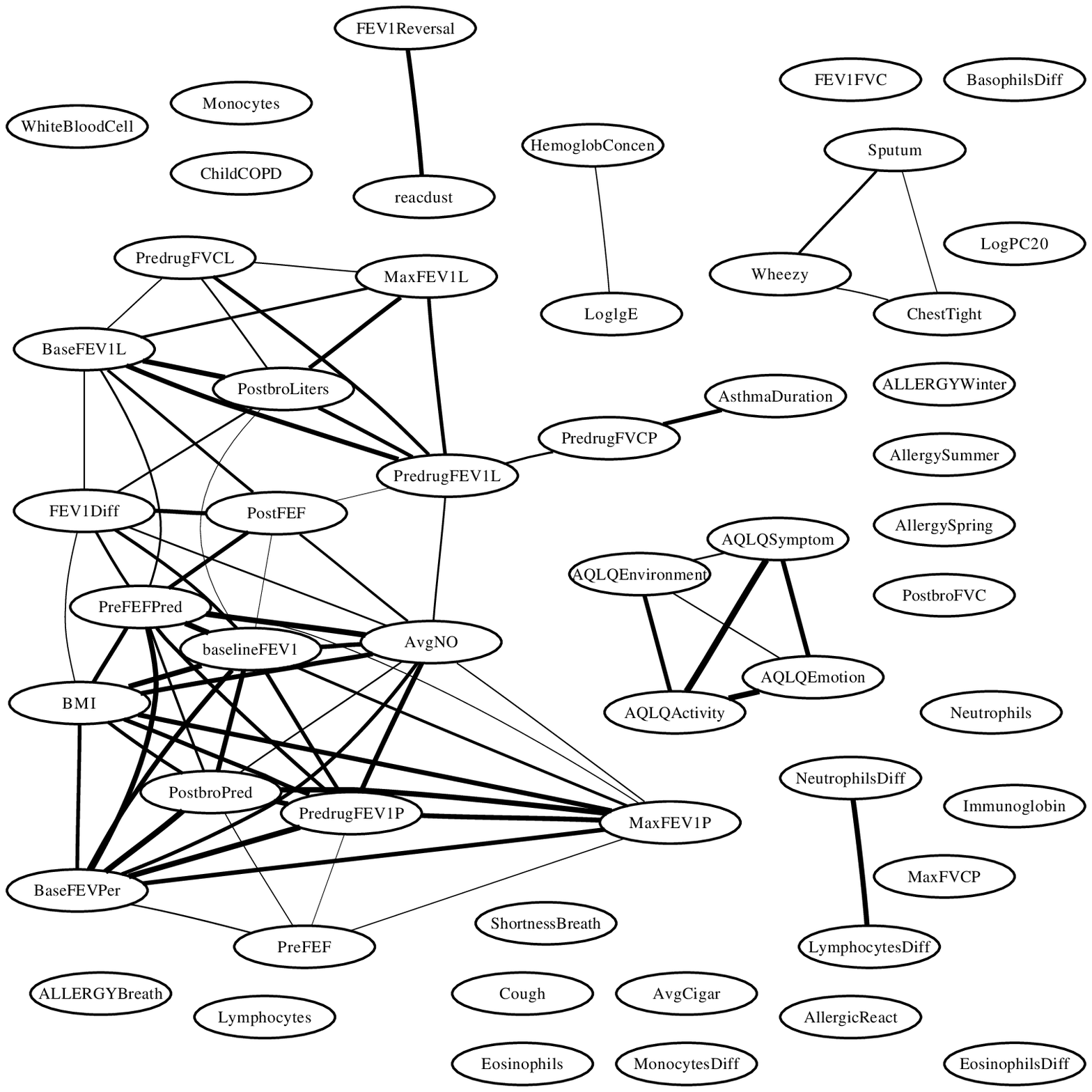}}
\end{picture}
\caption{ Illustration of association analysis using phenotype correlation graph
for asthma dataset }
\label{fig:sarp_g}
\end{figure}

Recent advances in high-throughput genotyping technologies have
significantly reduced the cost and time of genome-wide screening of
individual genetic differences over millions of single nucleotide
polymorphism (SNP) marker loci, shedding light to an era of
``personalized genome"~\cite{hapmap:2005, wtccc:2007}. Accompanying this trend, clinical
and molecular phenotypes are being measured at phenome and
transcriptome scale over a wide spectrum of diseases in various
patient populations and laboratory models, creating an imminent need
for appropriate methodology to identify omic-wide association
between genetic markers and complex traits which are implicative of
causal relationships between them. Many statistical approaches have
been proposed to address various challenges in identifying genetic
locus associated with the phenotype from a large set of markers,
with the primary focus on problems involving a univariate
trait~\cite{Li:2007, Ser:2007, Malo:2008}. However, in modern
studies the patient cohorts are routinely surveyed with a large
number of traits (from measures of hundreds of clinical phenotypes to
genome-wide profiling of thousands of gene expressions), many of
which are correlated among them. For example, in Figure
\ref{fig:sarp_g}, the correlation structure of the 53 clinical
traits in the asthma dataset collected as a part of the Severe
Asthma Research Program (SARP)~\cite{sarp:2007} is represented as a network,
with each trait as a node, the interaction between two traits as an
edge, and the thickness of an edge representing the strength of
correlation. Within this network, there exists several subnetworks
involving a subset of traits, and furthermore, the large subnetwork
on the left-hand side of Figure \ref{fig:sarp_g} contains two
subgroups of densely connected traits with thick edges. In order to
understand how genetic variations in asthma patients affect various
asthma-related clinical traits in the presence of such a complex
correlation pattern among phenotypes, it is necessary to consider
all of the traits jointly and take into account their correlation
structure in the association analysis.
Although numerous research efforts have been devoted to studying the
interaction patterns among many quantitative traits represented as
networks ~\cite{Butte:2006, Mehan:2008, Toh:2002, Fried:2004,
Segal:2003, Wille:2004, Scha:2005, Basso:2005} as well as
discovering network submodules from such networks~\cite{Segal:2003,
Hu:2005}, this type of network structure has not been exploited in
association mapping~\cite{Str:2005, Cheung:2005}. Many of the
previous approaches examined one phenotype at a time to localize the
SNP markers with a significant association and combined the results
from a set of such single-phenotype association mapping across
phenotypes. However, we conjecture that one can detect additional
weak associations and at the same time reduce false signals 
by combining the information across multiple phenotypes
under a single statistical framework.


In QTL mapping studies with pedigree data, a number of approaches
have been proposed to detect pleiotropic effect of markers on
multiple correlated traits by considering the traits jointly.
However, these approaches involve only a weak and indirect form of
structural information present in the phenotypes. The methods based
on multivariate regression~\cite{Knott:2000, Xu:2005, Liu:2007} with
multiple outcomes were concerned with finding genetic loci that
influence all of the phenotypes jointly, rather than explicitly
taking into account the complex interaction patterns among the
phenotypes. A different approach has been proposed that first
applies principle component analysis (PCA) to the phenotypes and
uses the transformed phenotypes in a single-phenotype association
test~\cite{Wel:1996, Man:1998}. The transformation via PCA allows to
extract the components that explain the majority of variation in
phenotypes, but has a limitation in that it is not obvious how to
interpret the derived phenotypes.

More recently, in expression quantitative trait locus (eQTL)
analysis with microarray gene expression measurements treated as
quantitative traits, researchers have begun to combine an explicit
representation of correlation structure in phenotypes, such as gene
networks, with genotype information to search for genetic causes of
perturbations of a subset of highly correlated
phenotypes~\cite{koller:2006,schadt:mouse,schadt:decode,brem:2008}. A
module network~\cite{Segal:2003}, which is a statistical model
developed for uncovering regulatory modules from gene expression
data, was extended to incorporate genotypes of regulators, such that
the expression of genes regulated by the same regulator was
explained by the variation of both the expression level and the
genotype of the regulators~\cite{koller:2006}. Although the model was
able to identify previously unknown genetic perturbations in yeast
regulatory network, the genotype information used in the model was
limited to markers in regulators rather than the whole genome.
Several other studies incorporated a gene co-regulation network in a
genome-wide scan for association. In a network eQTL association
study for mouse~\cite{schadt:mouse}, a gene co-regulation network was
learned, a clustering algorithm was applied to this network to
identify subgroups of genes whose members participate in the same molecular
pathway or biological process, and then, a single-phenotype analysis
was performed between genotypes and the phenotypes within each
subgroup. If the majority of phenotypes in each subgroup were mapped
to the common locus in the genome, that locus was declared to be
significantly associated with the subgroup. Using this approach, new
obesity-related genes in mouse were identified by examining the
network module associated with the genetic locus previously
associated with obesity-related traits such as body mass index and
cholesterol level. A similar analysis was performed on yeast, where
clusters of yeast genes were mapped to a common eQTL
hotspots~\cite{brem:2008}. One of the main disadvantages of this
approach is that it first applies a clustering algorithm to identify
subgroups of phenotypes in the network, rather than directly
incorporating the network itself as a correlation structure, since
the full network contains much richer information about complex
interaction patterns than the clusters of phenotypes. Another
disadvantage of this approach is that it relies on a set of
single-phenotype statistical tests and combines the results
afterwards in order to determine whether a marker is significantly
affecting a subgroup of phenotypes, thus requiring a substantial effort
in conducting appropriate multiple hypothesis testing. We believe
that an approach that considers markers and all of the phenotypes
jointly in a single statistical method has the potential to increase
the power of detecting weak associations and reduce susceptibility
to noise.

\begin{figure}[t!]
\centering
\begin{tabular}{ccc}
\includegraphics[scale = 0.18,angle=-90, bb=130 150 520 780]{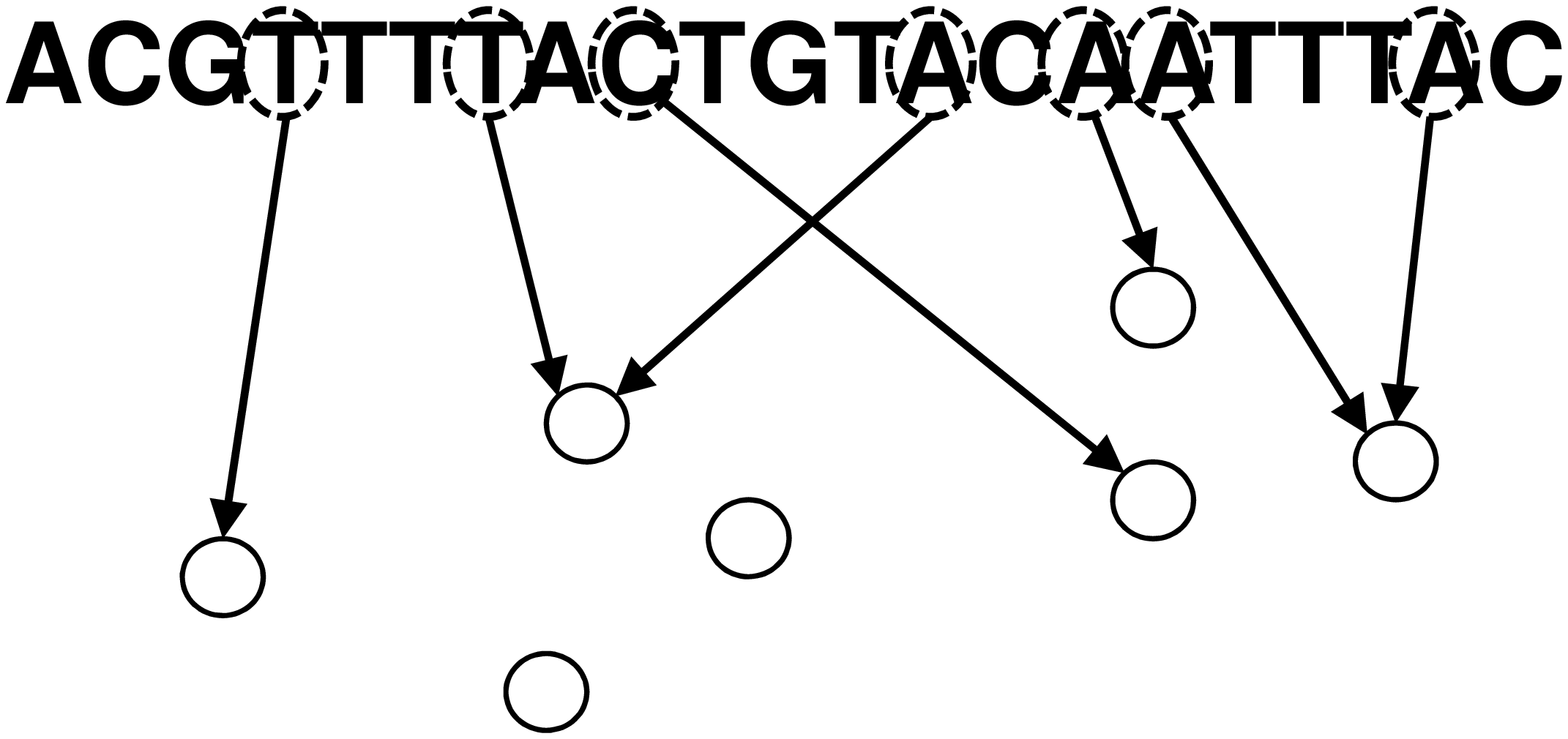} &
\includegraphics[scale = 0.18,angle=-90, bb=130 150 520 780]{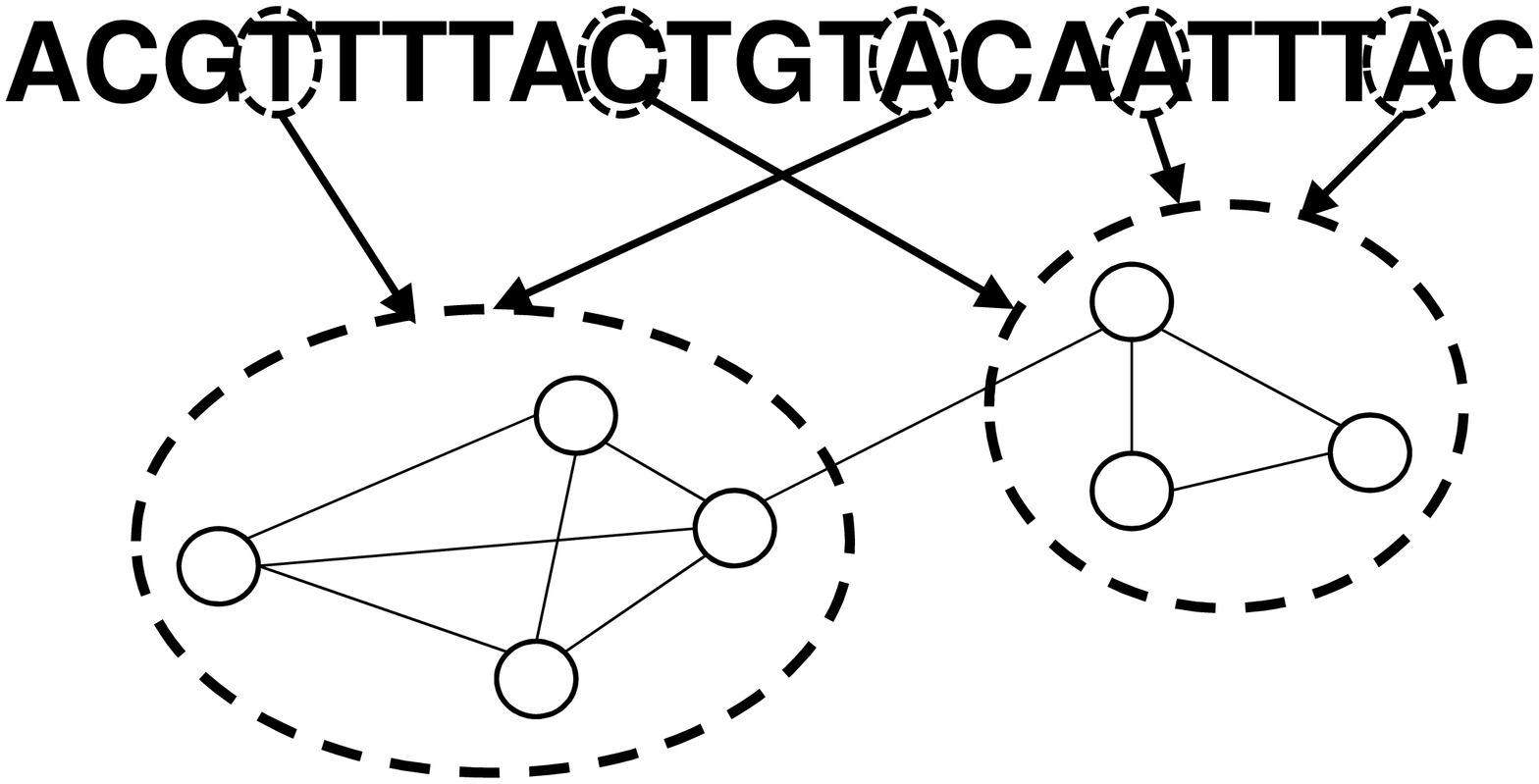} &
\includegraphics[scale = 0.18,angle=-90, bb=130 150 520 780]{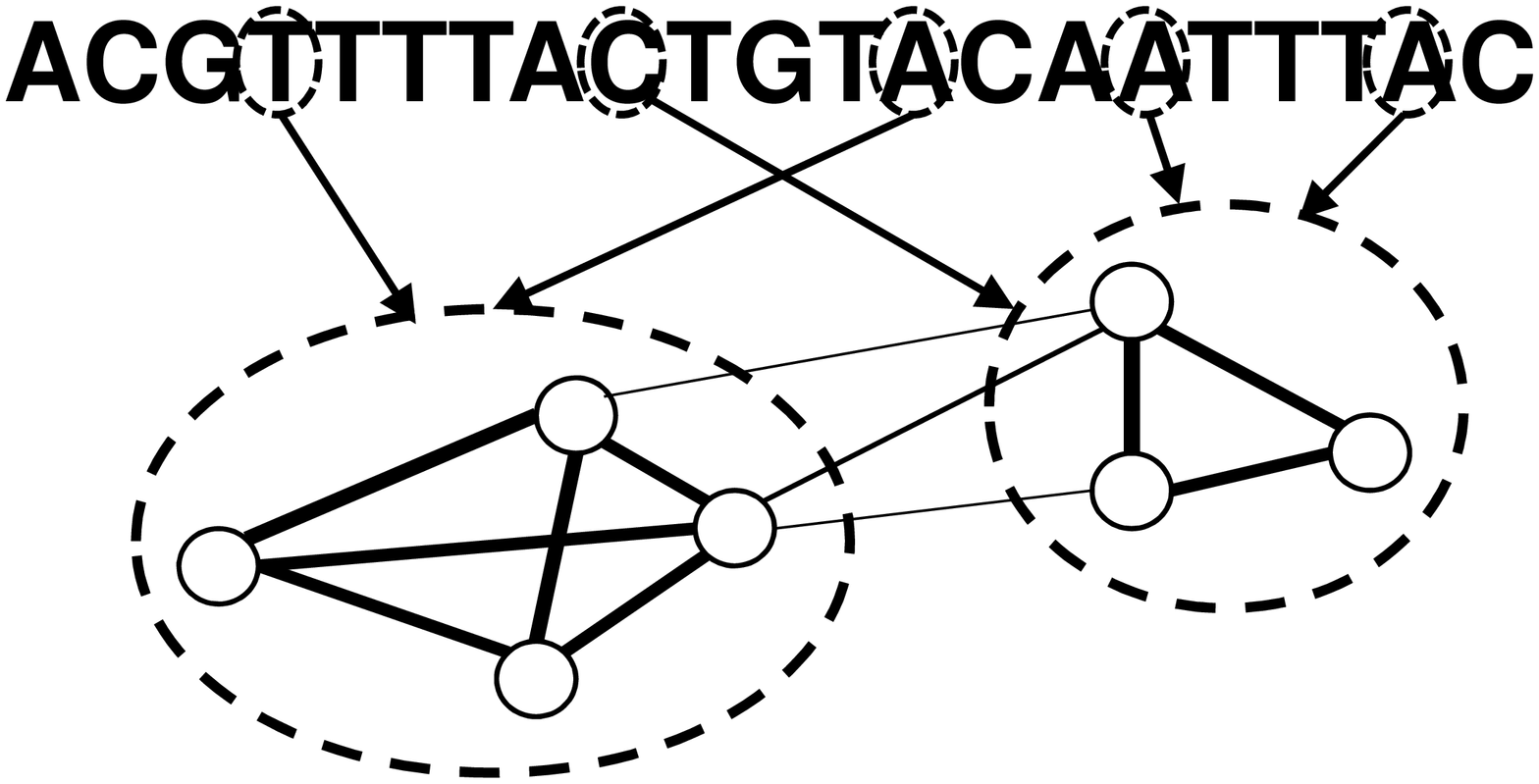} \\
A & B & C
\end{tabular}
\caption{Illustrations for multiple output regression with A: lasso, B:
graph-constrained fused lasso, and C: graph-weighted fused lasso.}
\label{fig:QTL}
\end{figure}

In this article, we propose a family of methods, called the {\it
graph-guided fused lasso} (GFlasso), that fully incorporates the
quantitative-trait network as an explicit representation for correlation
structure without applying additional clustering algorithms to
phenotypes. Our methods combine multiple phenotypes in a single
statistical framework, and analyze them jointly to identify SNPs
perturbing a subset of tighly correlated phenotypes instead of
combining results from multiple single-phenotype analyses. The
proposed methods leverage a dependency graph defined on multiple
quantitative traits such as the graph for the asthma-related traits
shown in Figure \ref{fig:sarp_g}, assuming that such a graph
structure is available from preprocessing steps or as prior
knowledge from previous studies.
It is reasonable to assume that when a subset of phenotypes are
highly correlated, the densely connected subgraphs over
these correlated traits contain variables that are more likely to
be synergistically influenced by the same or heavily overlapping
subset(s) of SNPs with similar strength than an arbitrary subset of
phenotypes.

The proposed approach is based on a multivariate regression
formalism with the $L_1$ penalty, commonly known as the lasso, that
achieves ``sparsistency" in the estimated model by setting many of
the regression coefficients for irrelevant markers to
zero~\cite{lasso,Zhao:2006}. As a brief digression for clarity, sparsistancy
refers to an asymptotic property in high-dimensional
statistical inference that for the estimator of a $p$-dimensional
vector $\vec\theta$ from $n$ iid samples, where $p$ can be $\gg n$,
the probability of recovering the true non-zero elements $S = \{i \
: \ \theta_i \neq 0\}$ in the estimator approach one in the limit, if
the true non-zero elements are sparse in the sense that $|S| \le n
\ll p$~\cite{Zhao:2006}. This property of the lasso makes it a natural approach
for genome wide association analysis, where the marker
genotypes are treated as the predictors, the phenotype in question
is treated as the response, and the (sparse) set of markers having
non-zero regression coefficients are interpreted as the markers
truely associated with the phenotype. However, when applied to
association mapping with multivariate traits, the lasso is
equivalent to a single-trait analysis that needs to be repeated over
every single trait. In other words, for a collection of traits, each trait
would be treated as independent of all other traits, and every trait
would be regressed on a common set of marker genotypes via its
own lasso (Figure \ref{fig:QTL}A), ignoring the possible coupling among
traits. Our innovations in GFlasso that enable a departure from the
baseline lasso for a single trait is that, in addition to the
lasso penalty, we employ a ``fusion penalty" that fuses regression
coefficients across correlated phenotypes, using either unweighed or
weighted connectivity of the phenotype graph as a guide. This
additional penalty will encourage sharing of common predictors
(i.e., associated markers) to coupled responses (i.e., traits). The
two fusion schemes lead to two variants of the GFlasso: a {\it
graph-constrained fused lasso} (${\rm G_cFlasso}$) based on only the
graph topology (Figure \ref{fig:QTL}B), and a {\it graph-weighted
fused lasso} (${\rm G_wFlasso}$) that offers a flexible range of
stringency of the graph constraints through edge weights (Figure~\ref{fig:QTL}C). 
We developed an efficient algorithm based on quadratic programming
combined with gradient search for estimating the regression
coefficients under GFlasso. The results on two datasets, one
simulated from HapMap SNP markers and the other collected from
asthma patients, show that our method outperforms competing
algorithms in identifying markers that are associated with a
correlated subset of phenotypes.

\section{Material and Methods}

\subsection{Lasso Regression for Multiple Independent Phenotypes}

Let $\mathbf{X}$ be an $N \times J$ matrix of genotypes for $N$ individuals
and $J$ SNPs, where each element $x_{ij}$ of $\mathbf{X}$ is assigned 0, 1, or 2
according to the number of minor alleles at the $j$-th locus
of the $i$-th individual. Let $\mathbf{Y}$ denote an $N \times K$
matrix of $K$ quantitative trait measurements over the same set
of individuals. We use $\mathbf{y}_k$ to denote the $k$-th column of
$\mathbf{Y}$.
A conventional single-trait association via linear regression model
can be applied to this multiple-trait setting by fitting the model to
$\mathbf{X}$ and each of the $K$ traits $\mathbf{y}_k$'s separately:
    \begin{eqnarray}
    \mathbf{y}_k &=& \mathbf{X}\bm{\beta}_k+\bm{\epsilon}_k,
    \quad \forall k=1,\ldots, K,
    \label{eq:m}
    \end{eqnarray}
where $\bm{\beta}_k$ is a $J$-vector of regression coefficients for the $k$-th
trait that can be used in a statistical test to detect SNP markers with significant
association, and $\bm{\epsilon}_k$ is a vector of $N$ independent error terms with
mean 0 and a constant variance. We center each column of $\mathbf{X}$ and $\mathbf{Y}$
such that $\sum_i y_{ik}=0$ and $\sum_i x_{ij}=0$,
and consider the model in Equation (\ref{eq:m}) without an intercept.
We obtain the estimates of $\mathbf{B} = \{\bm{\beta}_1, \ldots, \bm{\beta}_K\}$
by minimizing the resisual sum of squares:
    \begin{eqnarray}
    \hat{\mathbf{B}} &=& \textrm{argmin}
        \sum_k (\mathbf{y}_k - \mathbf{X}\bm{\beta}_k)^{T}
            \cdot (\mathbf{y}_k - \mathbf{X}\bm{\beta}_k).
    \label{eq:ols}
    \end{eqnarray}
In a typical genome-wide association mapping, one examines a large number of
marker loci with the goal of identifying the region
associated with the phenotypes and markers in that region. A straight-forward
application of the linear regression method in Equation (\ref{eq:ols}) to
association mapping with large $J$ can cause several problems such as
an unstable estimate of regression coefficients and a poor interpretability
due to many irrelevant markers with non-zero regression coefficients.
Sparse regression methods such as forward stepwise selection~\cite{Weis:1980},
ridge regression~\cite{Hoerl:1975, Malo:2008}, and lasso~\cite{lasso}
that select a subset of markers with true association
have been proposed to handle the situation with large $J$.
Forward stepwise selection method iteratively selects one relevant marker at a time
while trying to improve the model fit based on Equation (\ref{eq:ols}),
but it may not produce an optimal solution becuase of the greedy nature of the algorithm.
Ridge regression has an advantage of performing the selection in a
continuous space by penalizing the residual sum of
square in Equation (\ref{eq:ols}) with the $L_2$ norm of $\bm{\beta}_k$'s and shrinking
the regression coefficients toward zero, but it does not set the regression coefficients of
irrelevant markers to exactly zero.
We use the lasso that penalizes the residual sum of square
with the $L_1$ norm of regression coefficients and has the property of 
setting regression coefficients with weak association markers exactly to zero,
thus offering the advantages of both forward stepwise selection and ridge regression.
The lasso estimate of the regression coefficients can be obtained by solving the following:
    \begin{eqnarray}
    \hat{\mathbf{B}}^{\textrm{lasso}} &=& \textrm{argmin}
        \sum_k (\mathbf{y}_k - \mathbf{X}\bm{\beta}_k)^{T}
            \cdot (\mathbf{y}_k - \mathbf{X}\bm{\beta}_k)
        + \lambda \sum_{k,j} |\beta_{kj}|
    \label{eq:lasso}
    \end{eqnarray}
where $\lambda$ is a regularization parameter that controls the amount of sparsity
in the estimated regression coefficients. Setting $\lambda$ to a large value
increases the amount of penalization,
setting more regression coefficients to zero. Many fast algorithms are available
for solving Equation (\ref{eq:lasso})~\cite{lasso, lars}.

The lasso for multiple-trait association mapping in Equation
(\ref{eq:lasso}) is equivalent to solving a set of $K$ independent
regressions for each trait with its own $L_1$ penalty, and
does not provide a mechanism to combine information across
multiple traits such that the estimates reflect the potential
relatedness in the regression coefficients for those correlated
traits that are influenced by common SNPs. 
However, several traits are often highly
correlated such as in gene expression of co-regulated genes in eQTL
study, and there might be genotype markers that are jointly
associated with those correlated traits. 
Below, we extend the
standard lasso and propose a new penalized regression method for
detecting markers with pleiotropic effect on correlated quantitative traits.

\subsection{Graph-Guided Fused Lasso for Multiple Correlated Phenotypes}

In order to identify markers that are predictive of multiple
phenotypes jointly, we represent the correlation structure over the
set of $K$ traits as an edge-weighted graph, and use this graph to guide
the estimation process of the regression coefficients within the
lasso framework. We assume that we have available from a
pre-processing step a phenotype correlation graph $G$ consisting of
a set of nodes $V$, each representing one of the $K$ traits, and a
set of edges $E$. In this article, we adopt a simple and
commonly-used approach for learning such graphs, where we first
compute pairwise Pearson correlation coefficients for all pairs of
phenotypes using $\mathbf{y}_k$'s, and then connect two nodes with
an edge if their correlation coefficient is above the given
threshold $\rho$. We set the weight of each edge $(m,l) \in E$ to
the absolute value of correlation coefficient $|r_{m,l}|$, so that
the edge weight represents the strength of correlation between the
two nodes. This thresholded correlation graph is also known as a
relevance network, and has been widely used as a representation of
gene interaction networks ~\cite{Butt:2000, Cart:2004}. It is worth
pointing out that the choice of methods for obtaining the phenotype
network is not a central issue of our method. Other variations of
the standard relevance network have been suggested~\cite{Zhang:2005},
and any of these graphs can also be used within our proposed
regression methods. Below, we first introduce \Gc~ that makes use of
unweighted graph, and further extend this method to \Gwa~to take
into account the full information in the graph including edge
weights.

Given the correlation graph of phenotypes, it is reasonable to
assume that if two traits are highly correlated and connected
with an edge in the graph, their variation across individuals
might be explained by genetic variations at the same loci,
possibly having the same amount of influence on each trait.
In \Gc, this assumption is expressed as an additional penalty term
that fuses two regression
coefficients $\beta_{jm}$ and $\beta_{jl}$ for each marker $j$
if traits $m$ and $l$ are connected with an edge in the graph, as follows:
    \begin{eqnarray}
    \hat{\mathbf{B}}^{\textrm{GC}} = \textrm{argmin}  \
        \sum_k (\mathbf{y}_k - \mathbf{X}\bm{\beta}_k)^{T}
            \cdot (\mathbf{y}_k - \mathbf{X}\bm{\beta}_k)
    \quad\quad\quad\quad\quad\nonumber \\
    \quad\quad\quad
    + \lambda \sum_k \sum_j |\beta_{jk}| +
        \gamma \sum_{(m,l)\in E} \sum_j |\beta_{jm}-\textrm{sign}(r_{ml})\beta_{jl}|,
        \label{eq:gc_lasso}
    \end{eqnarray}
where $\lambda$ and $\gamma$ are regularization parameters that
determine the amount of penalization. The last term in Equation
(\ref{eq:gc_lasso}) is called a fusion penalty~\cite{flasso}, and
encourages $\beta_{jm}$ and $\textrm{sign}(r_{m,l})\beta_{jl}$ to
take the same value by shrinking the difference between them toward
zero. A larger value for $\gamma$ leads to a greater fusion effect, or
greater sparsity in $|\beta_{jm}-\mysgn(r_{m,l})\beta_{jl}|$'s. We
assume that if two traits $m$ and $l$ connected with an edge in $G$
are negatively correlated with $r_{ml}<0$, the effect of a common
marker on those traits takes an opposite direction, and we fuse
$\beta_{jm}$ and $(-\beta_{jl})$, or equivalently, $\beta_{jm}$ and
$\textrm{sign}(r_{m,l})\beta_{jl}$. When the fusion penalty is
combined with the lasso penalty as in Equation (\ref{eq:gc_lasso}),
the lasso penalty sets many of the regression coefficients to zero,
and for the remaining non-zero regression coefficients, the fusion
penalty flattens the values across multiple highly correlated
phenotypes for each marker so that the strength of influence of each
marker becomes similar across those correlated traits. The idea of
fusion penalty has been first used in the classical regression
problem over univariate response (i.e., single-output) from
high-dimensional covariates to fuse the regression coefficients of
two adjacent covariates when the covariates are assumed to be
ordered such as in time~\cite{flasso}. This corresponds to coupling
pairs of elements in the adjacent rows of the same column in the
coefficient matrix $\mathbf{B}$ in Equation (\ref{eq:gc_lasso}). In
\Gc, we employ a similar strategy in a multiple-output regression in
order to identify pleiotropic effect of markers, and let the trait
correlation graph determine which pairs of regression coefficients
should be fused. Now, every such coupled coefficient pair corresponds
to the elements of the corresponding two columns in the same row of
matrix $\mathbf{B}$ in Equation (\ref{eq:gc_lasso}).
It is possible to show the asymptotic properties of estimators 
of the GFlasso methods as $N \to \infty$ analogous to the ones previously shown 
for the lasso and the fused lasso~\cite{flasso, Knight:2000}.

In a multiple-trait association mapping, networks of clinical traits or
molecular traits (i.e., gene expressions) typically contain many
subnetworks within which nodes are densely connected, and we are
interested in finding the genetic variants that perturb the entire
set of traits in each
subnetwork. This can potentially increase the power of detecting
weak associations between genotype and phenotype that may be
missed when each phenotype is considered independently. When used in
this setting, the \Gc~ looks for associations between a
genetic marker and a subgraph of phenotype network rather than
a single phenotype.
Unlike other previous approaches for detecting pleiotropic effect
that first apply clustering algorithms to learn subgroups of traits
and then search for genetic variations that perturb the subgroup,
\Gc~ uses the full information on correlation structure
in phenotypes available as a graph, where the subgroup information
is embedded implicitly within the graph as densely connected
subgraphs. Although the fusion penalty in the \Gc~
is applied locally to a pair of regression coefficients for neighboring trait pairs
in the graph, this fusion effect propagates to the regression coefficients for
other traits that are connected to them in the graph.
For densely connected nodes, the fusion is effectively
applied to all of the members of the subgroup, and the set of
non-zero regression coefficients tend to show a block structure with
the same values across the correlated traits given a genotype
marker with pleiotropic effect on those traits, as we demonstrate in experiments.
If the edge connections are sparse within a group of nodes, the corresponding
traits are only weakly related, and there is little propagation of fusion effect
through edges in the subgroup.
Thus, the \Gc~ incorporates the subgrouping information through
the trait correlation graph in a more flexible manner
compared to previous approaches.

Now, we present a further generalization of \Gc~ that exploit the full
information in the phenotype networks for association mapping. Note
that the only structural information used in the \Gc~ is the
presence or absence of edges between two phenotypes in the graph.
The \Gwa~ is a natural extension of the \Gc~ that takes into account
the edge weights in graph $G$ in addition to the graph topology.
The \Gwa~ weights each term in the fusion penalty in Equation (\ref{eq:gc_lasso}) 
by the amount of correlation between the two phenotypes
being fused, so that the amount of correlation controls the amount
of fusion. More generally, \Gwa~ weights each term in the fusion
constraint in Equation (\ref{eq:gc_lasso}) with a monotonicaly
increasing function of the absolute values of correlations, and
finds an estimate of the regression coefficients as follows:
    \begin{eqnarray}
    \hat{\mathbf{B}}^{\textrm{GW}} = \textrm{argmin} \
        \sum_k (\mathbf{y}_k - \mathbf{X}\bm{\beta}_k)^{T}
            \cdot (\mathbf{y}_k - \mathbf{X}\bm{\beta}_k)
    \quad\quad\quad\quad\quad\quad\quad\quad \nonumber \\
    \quad\quad\quad\quad\quad\quad
        + \lambda \sum_k \sum_j |\beta_{jk}|
        + \gamma \sum_{(m,l)\in E} f(r_{ml}) \sum_j
        |\beta_{jm}-\textrm{sign}(r_{ml})\beta_{jl}|.
        \label{eq:gw_lasso}
    \end{eqnarray}
If the two phenotypes $m$ and $l$ are highly correlated in graph $G$
with a relatively large edge weight, the fusing effect increases
between these two phenotypes since the difference between the two
corresponding regression coefficients $\beta_{jm}$ and $\beta_{jl}$
is penalized more than for other pairs of phenotypes with weaker correlation.
In this article, we consider $f_1(r)=|r|$ for
${\rm G^1_wFlasso}$ and $f_2(r) = r^2$ for ${\rm G^2_wFlasso}$. We
note that the ${\rm G_cFlasso}$ is a special case of the ${\rm
G_wFlasso}$ with $f(r)=1$.

The optimization problems in Equations (\ref{eq:gc_lasso}) and
(\ref{eq:gw_lasso}) can be formulated as a quadratic programming as
described in Appendix, and there are many publicly available
software packages that efficiently solve such quadratic programming
problems. The regularization parameters $\lambda$ and $\gamma$ can
be determined by a cross-validation or a validation set, although for a large problem, a
grid search for the $\lambda$ and $\gamma$ can be time-consuming. In
order to improve the efficiency in determining the $\lambda$ and
$\gamma$, we use a gradient-descent type of algorithm as described
in Appendix.


\subsection{Simulation Scheme for Model Validation}

We simulated genotype data for 250 individuals based on the HapMap data
in the region of 8.79-9.20M in chromosome 7.
The first 60 individuals of the genotype data came from the parents of the HapMap
CEU panel. We generated genotypes for additional 190 individuals by randomly
mating the original
60 individuals on the CEU panel. We included only those SNPs with
minor allele frequency greater than 0.1.
Since our primary goal is to measure the performance of the association
methods in the case of multiple correlated phenotypes, we
sampled 50 SNPs randomly from the 697 SNPs in the region in order to
reduce the correlation among SNPs from the linkage disequilibrium.

Given the simulated genotype, we generated the true associations
represented as regression coefficients $\mathbf{B}$ and phenotype data as follows.
We assumed that the number of phenotypes is 10, and that there are
three groups of correlated phenotypes of size 3, 3, and 4, respectively,
so that the phenotypes in each group form a subnetwork in the
correlation graph of phenotypes.
We randomly selected three SNPs as affecting all of the phenotypes in
the first subnetwork, and four SNPs as influencing each of the remaining
two subnetwork. We assumed that there is one additional SNP
affecting phenotypes in the first two subnetworks, which corresponds to the
case of a SNP perturbing a super-network consisting of two subnetworks
such as the large subnetwork on the left-hand side of Figure \ref{fig:sarp_g}.
In addition, we assumed one additional SNP affecting all of the phenotypes.
We set the effect size of all of the true association SNPs to the
same value. Once we set the regression coefficients according to
this setup, we generated the phenotype data with noise distributed
as $N(0,1)$, using the simulated genotypes as covariates.

\subsection{Asthma Dataset}

We apply our methods to data collected from 543 asthma patients
as a part of the Severe Asthma Research Program (SARP).
The genotype data were obtained for
34 SNPs within or near IL-4R gene that spans a 40kb region on
chromosome 16. This gene has been previously shown to be implicated
in severe asthma~\cite{Wenz:2007}.
We used the publicly available software {\it PHASE}~\cite{Li:2003} to impute missing
alleles and phase the genotypes. The phenotype data included
53 clinical traits related to severe asthma such as age of onset, family history, and
severity of various symptoms. The phenotype correlation graph thresholded
at 0.7 as shown in Figure \ref{fig:sarp_g} reveals several subnetworks
of correlated traits. For example, the subset of traits related to lung physiology
(the nodes for baselineFEV1, PreFEFPred, PostbroPred, PredrugFEV1P, Max FEV1P, etc.)
form a large subnetwork on the left-hand side of Figure \ref{fig:sarp_g},
whereas traits representing quality of life of the patients (the nodes for AQLQ Environment,
AQLQSymptom, AQLQ Emotion, and AQLQ Activity) are found in a small subnetwork
near the center of Figure \ref{fig:sarp_g}. Our goal is to examine whether any of
the SNPs in the IL-4R gene are associated with a subnetwork of correlated traits
rather than an individual trait. We standardized measurements for each phenotype
to have mean 0 and standard deviation 1 so
that their values are roughly in the same range across phenotypes.

\section{Results}

We compare the results from our proposed methods, \Gc~, \Gw~ and \Gww,
with the ones from the single-marker analysis as well as multivariate
regression methods such as the ridge regression and the lasso
that do not use any structural information in the phenotypes.
In the ridge regression, we set the
regularization parameter to 0.0001, which is equivalent to adding
a small value of 0.0001 to the diagonal of $\mathbf{X}'\mathbf{X}$
to make the standard regression problem non-singular.
For the lasso and our proposed methods, we used the gradient-descent
search for the regularization parameters $\lambda$ and $\gamma$
as described in Appendix.
We used (-$\log$($p$-value)) for the standard single-marker
analysis, and the absolute value of regression coefficients $|\beta_{jk}|$'s
for the multivariate regression methods and our proposed methods, as a measure
of the strength of association.

\subsection{Simulation Study}

\begin{figure}[t!]\centering
\begin{tabular}{@{}c@{}c@{}c@{}}
\includegraphics[scale = 0.27,angle=0]{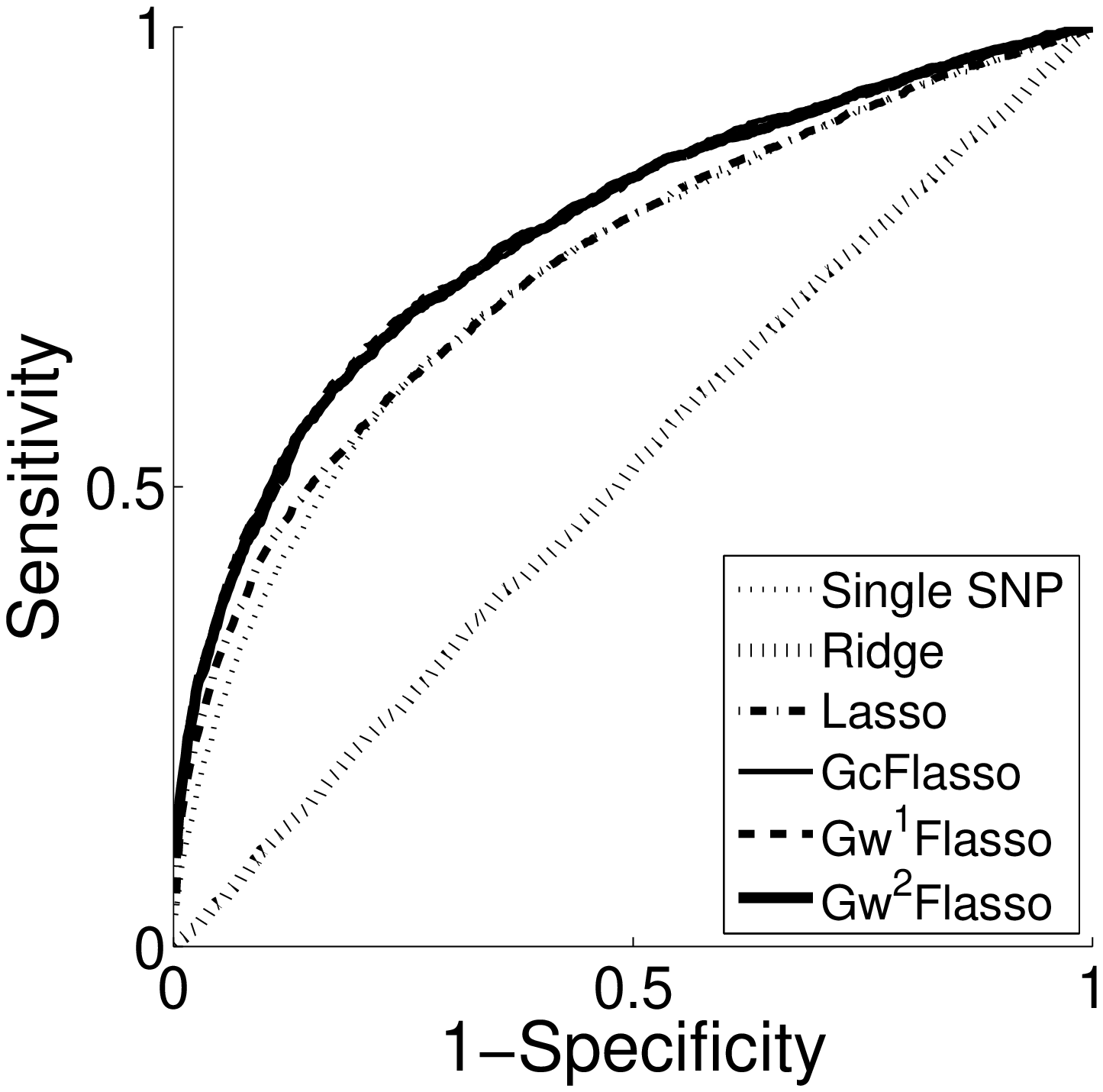}  &
\includegraphics[scale = 0.27,angle=0]{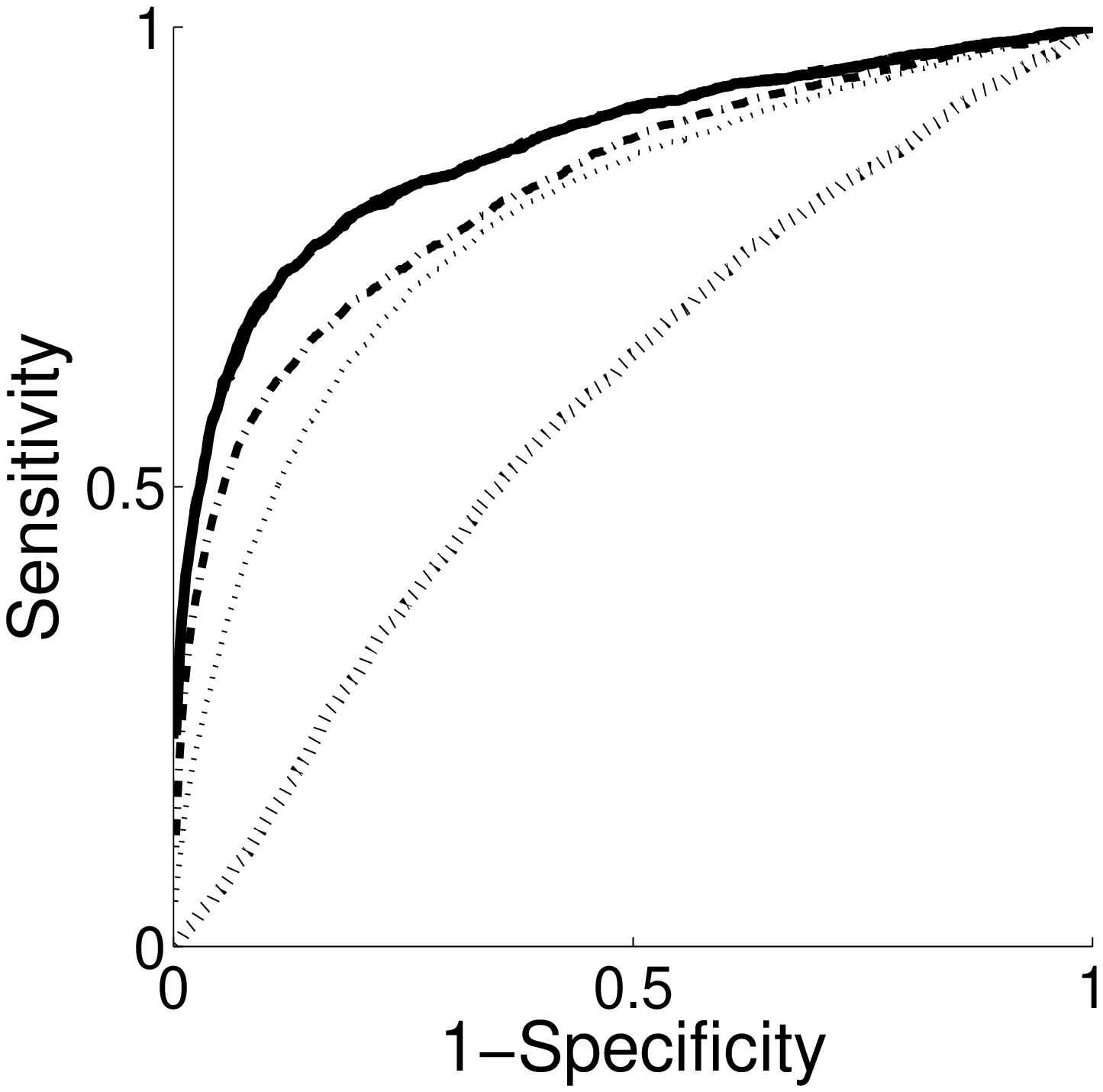}  &
\includegraphics[scale = 0.27,angle=0]{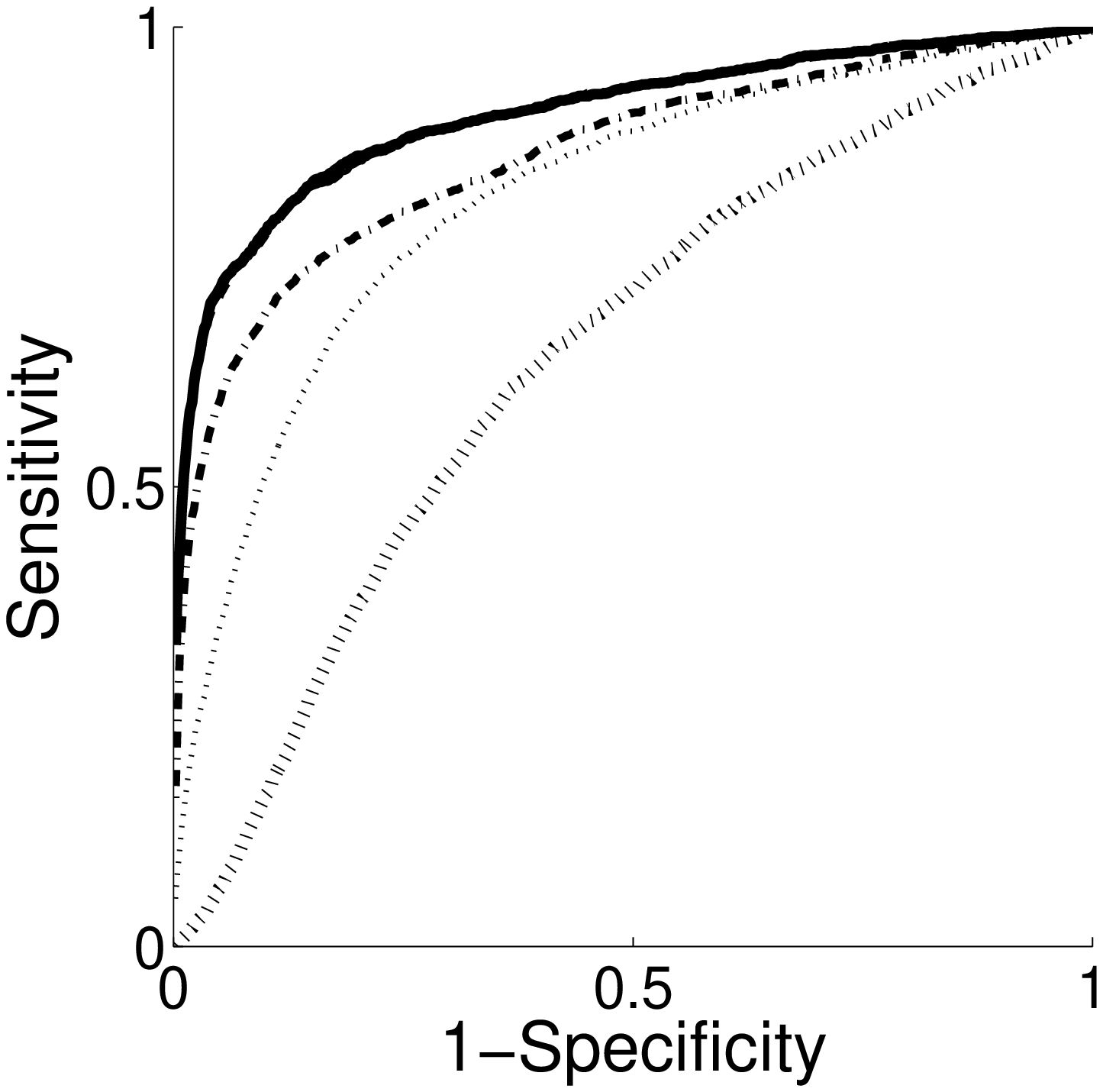}  \\
A & B & C
\vspace{3pt}
\\
\includegraphics[scale = 0.27,angle=0]{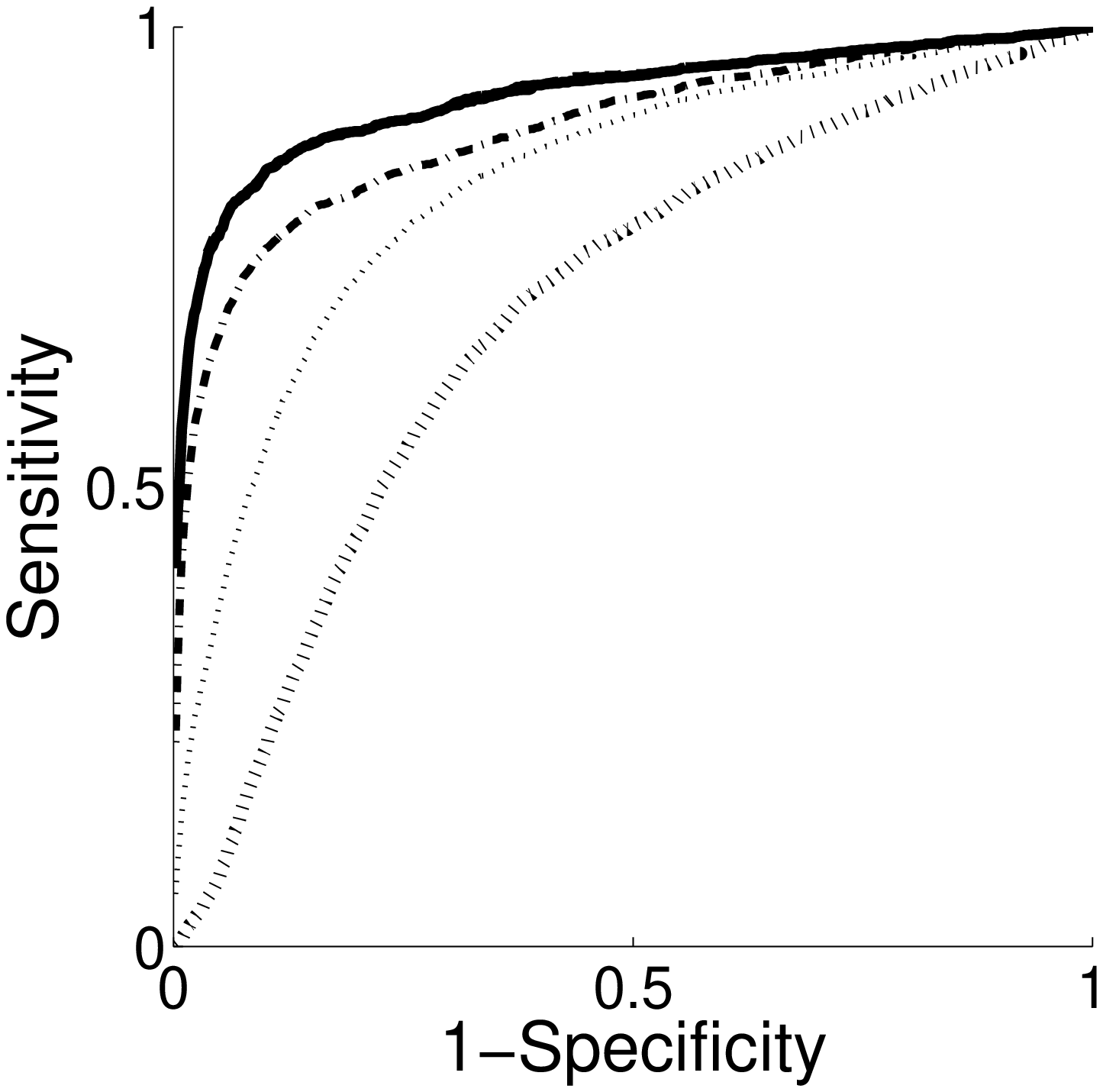}  &
\includegraphics[scale = 0.27,angle=0]{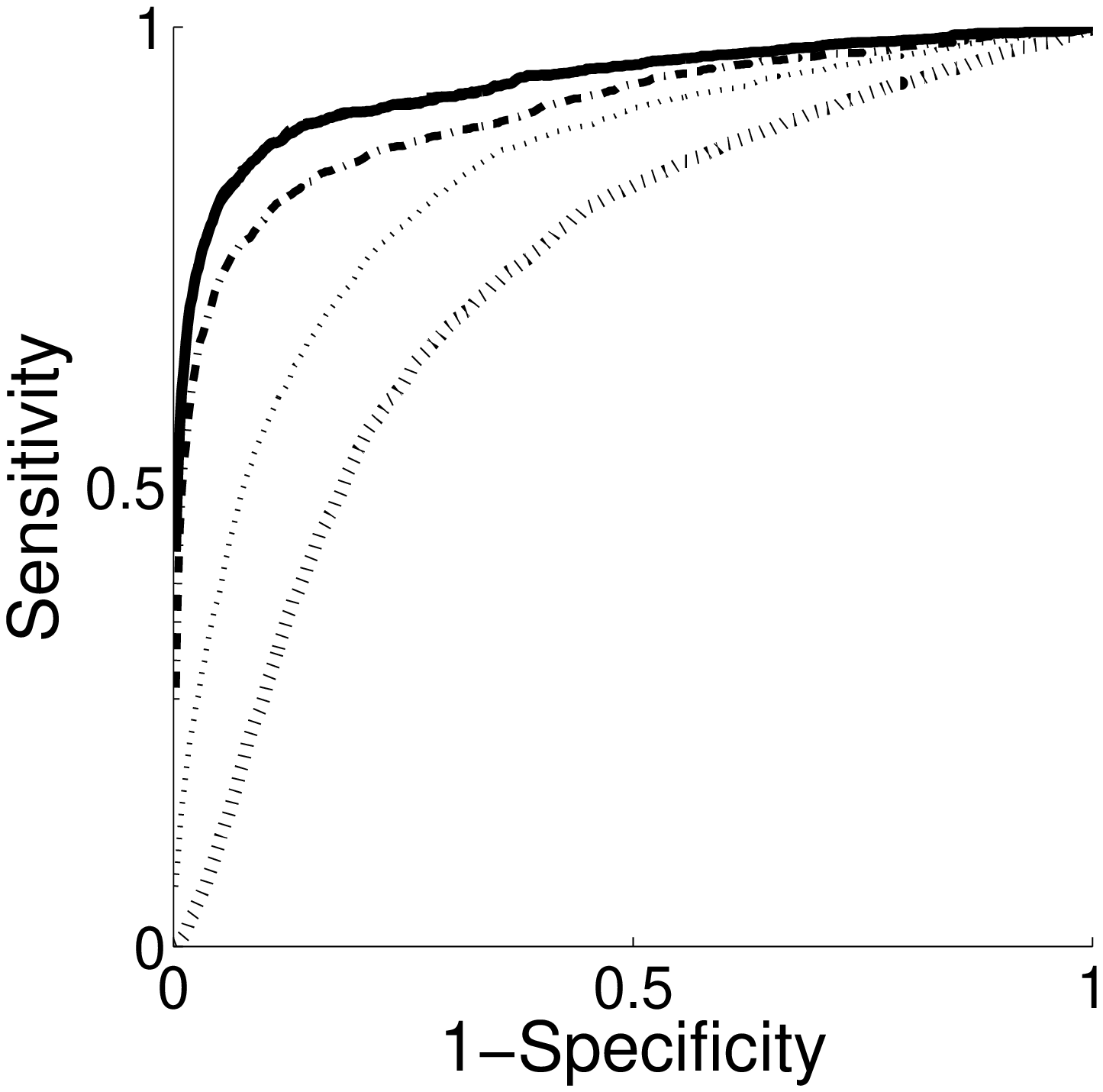}  & \\
D & E &
\\
\end{tabular}
\caption{ ROC curves for comparison of association
analysis methods with different sample size $N$.
A: $N$=50, B: $N$=100, C: $N$=150, D: $N$=200, and E: $N$=250. The effect
size is 0.5, and the threshold $\rho$ for the phenotype network is
set to 0.3. Note that the curves for the \Gc, \Gw, and \Gww~ almost
entirely overlap.
} \label{fig:sim_ss}
\end{figure}

We evaluate the performance of the association methods
based on two criteria, sensitivity/specificity and phenotype prediction error.
The sensitivity and specificity measure whether the given
method can successfully detect the true association SNPs with few false
positives. The (1-specificity) and sensitivity
are equivalent to type I error rate and (1-type II error rate),
and their plot is widely known as a receiver operating characteristic (ROC) curve.
The phenotype prediction error represents how accurately
we can predict the values of phenotypes given the genotypes
of new individuals, using the regression coefficients estimated
from the previously available genotype and phenotype data.
We generate additional dataset of 50 individuals, $\mathbf{y}^{\textrm{new}}$ and
$\mathbf{X}^{\textrm{new}}$, and compute the phenotype prediction error
as sum of squared differences between the true values 
$\mathbf{y}^{\textrm{new}}$ and predicted values 
$\hat{\mathbf{y}}^{\textrm{new}}$ of the phenotypes,
$\sum_k (\mathbf{y}_k^{\textrm{new}}-\hat{\mathbf{y}}^{\textrm{new}})'
\cdot (\mathbf{y}_k^{\textrm{new}}-\hat{\mathbf{y}}^{\textrm{new}})$,
where $\hat{\mathbf{y}_k}^{\textrm{new}}=\mathbf{X}^{\textrm{new}}\hat{\bm{\beta}}_k$.
For both criteria for measuring performance, we show results averaged over 50 randomly
generated datasets.

\begin{figure}[t!]\centering
\begin{tabular}{@{}c@{}c@{}c@{}c}
\includegraphics[scale = 0.27]{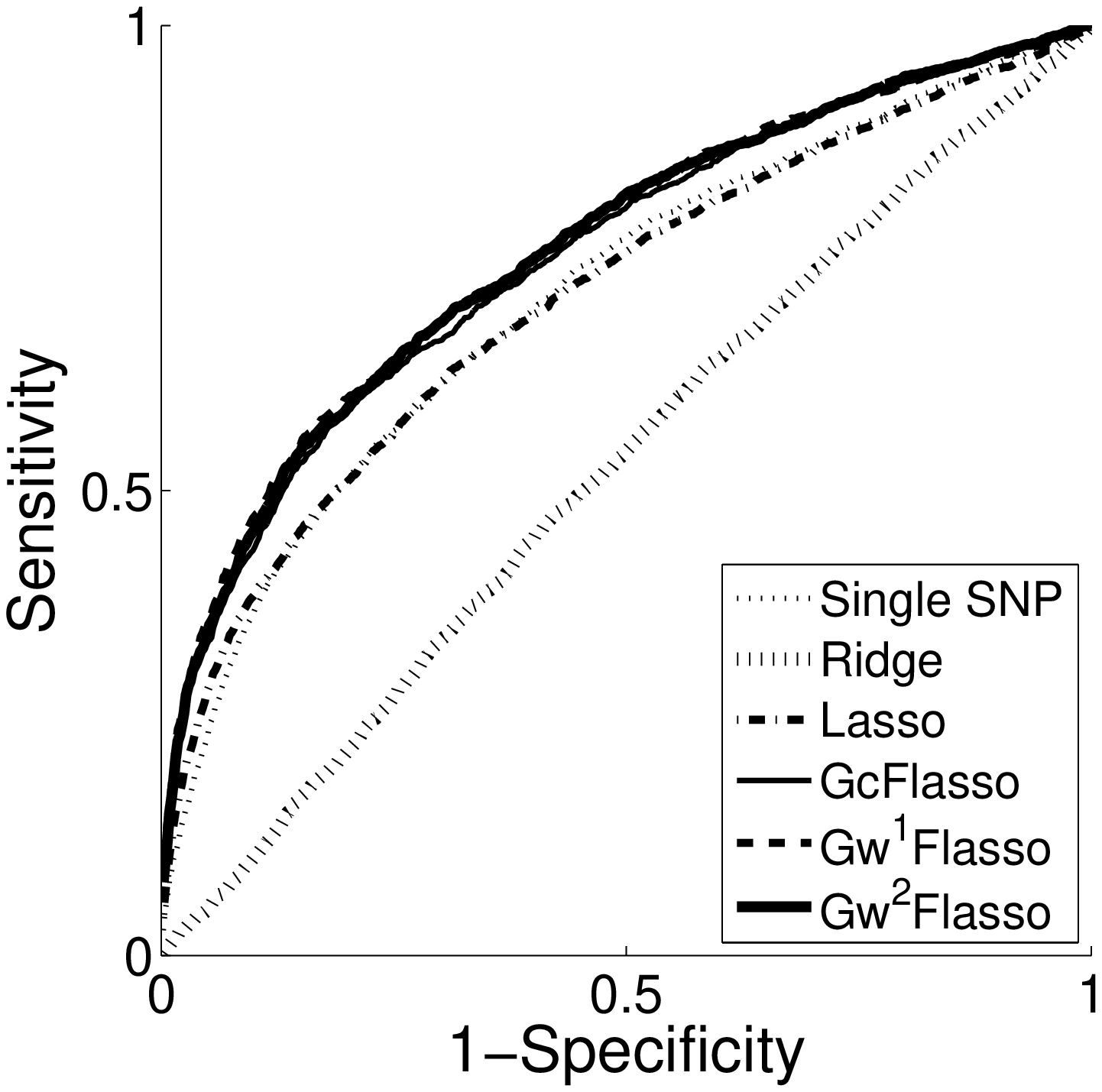} &
\includegraphics[scale = 0.27]{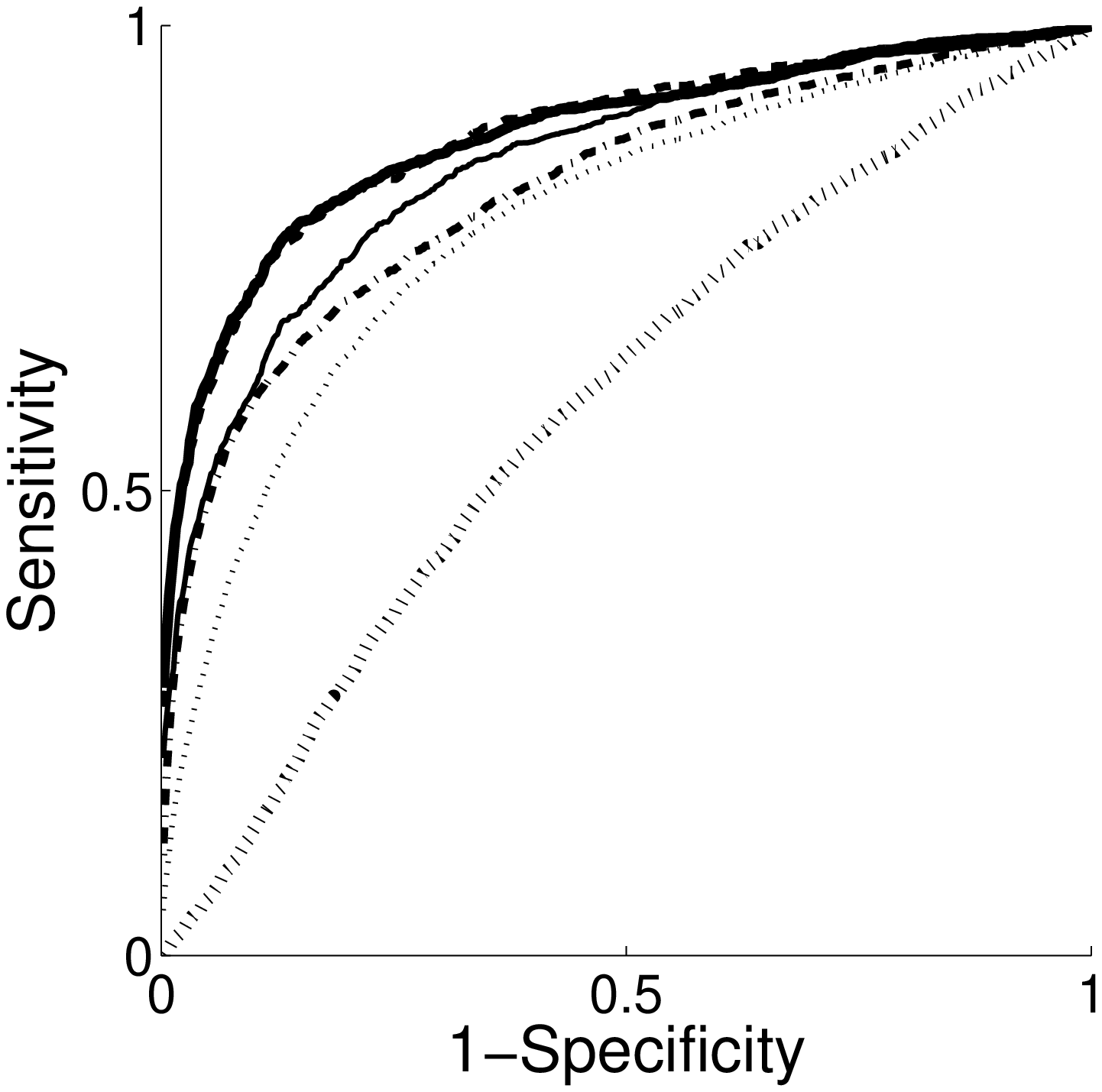} &
\includegraphics[scale = 0.27]{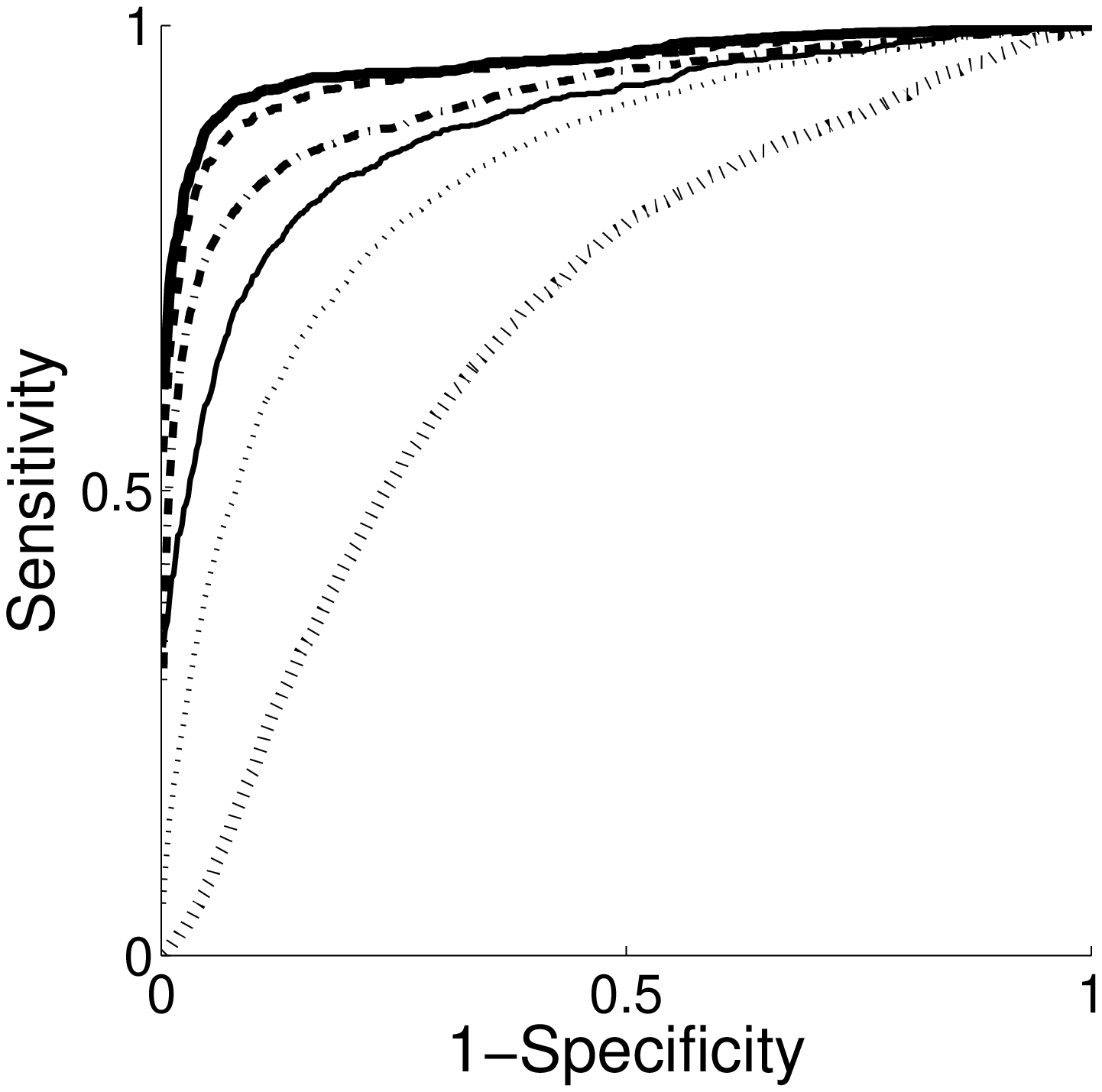} &
\includegraphics[scale = 0.27]{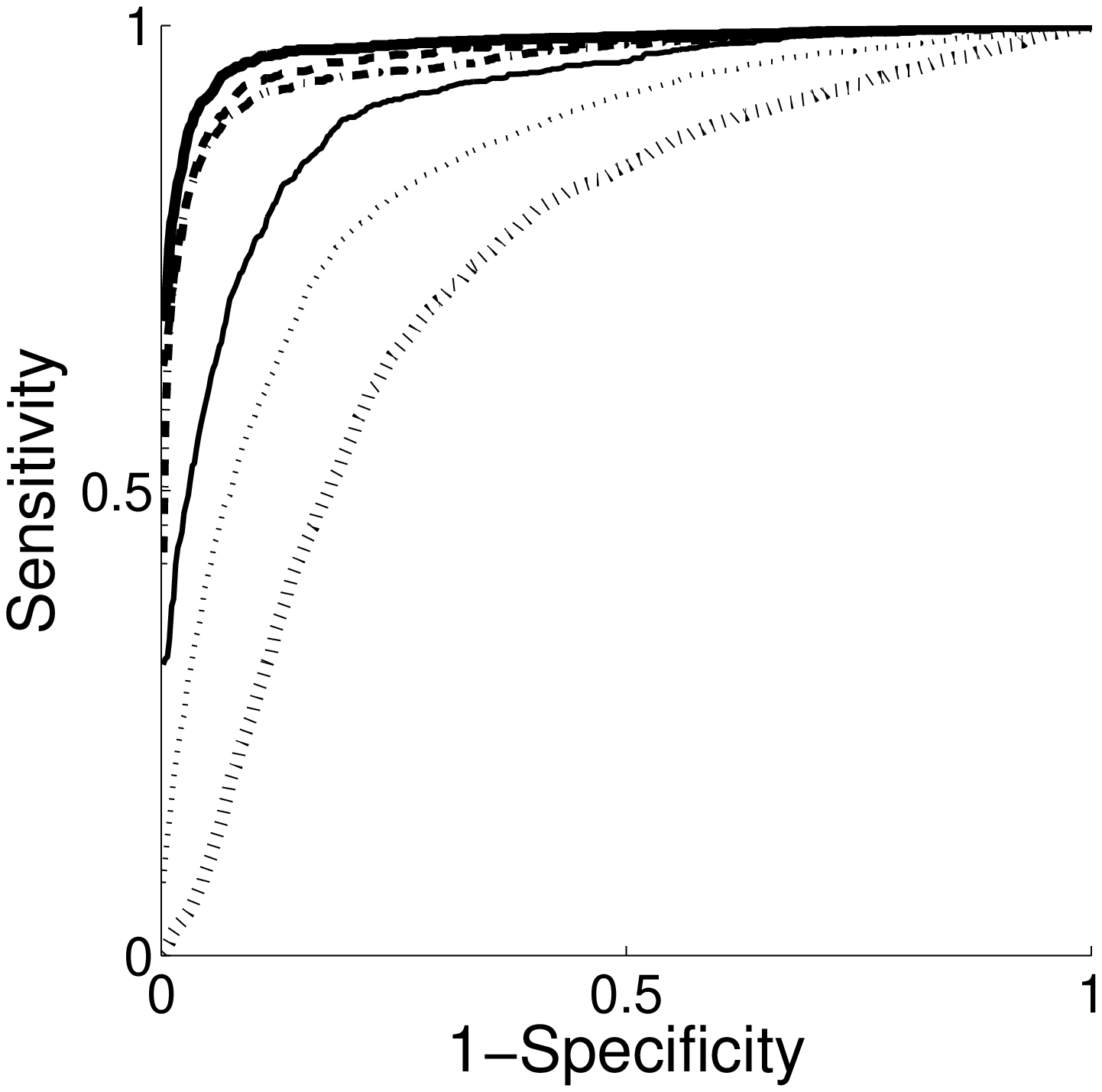} \\
A & B & C & D
\end{tabular}
\caption{ ROC curves for comparison of
association analysis methods with varying effect
size. Effect size is A: 0.3, B: 0.5, C: 0.8,
and D: 1.0. The sample size is 100, and
the threshold $\rho$ for the phenotype correlation graph
is 0.1.
} \label{fig:sim_b}
\end{figure}

\begin{figure}[t!]\centering
\begin{tabular}{@{}c@{}c@{}c@{}c}
\includegraphics[scale = 0.27]{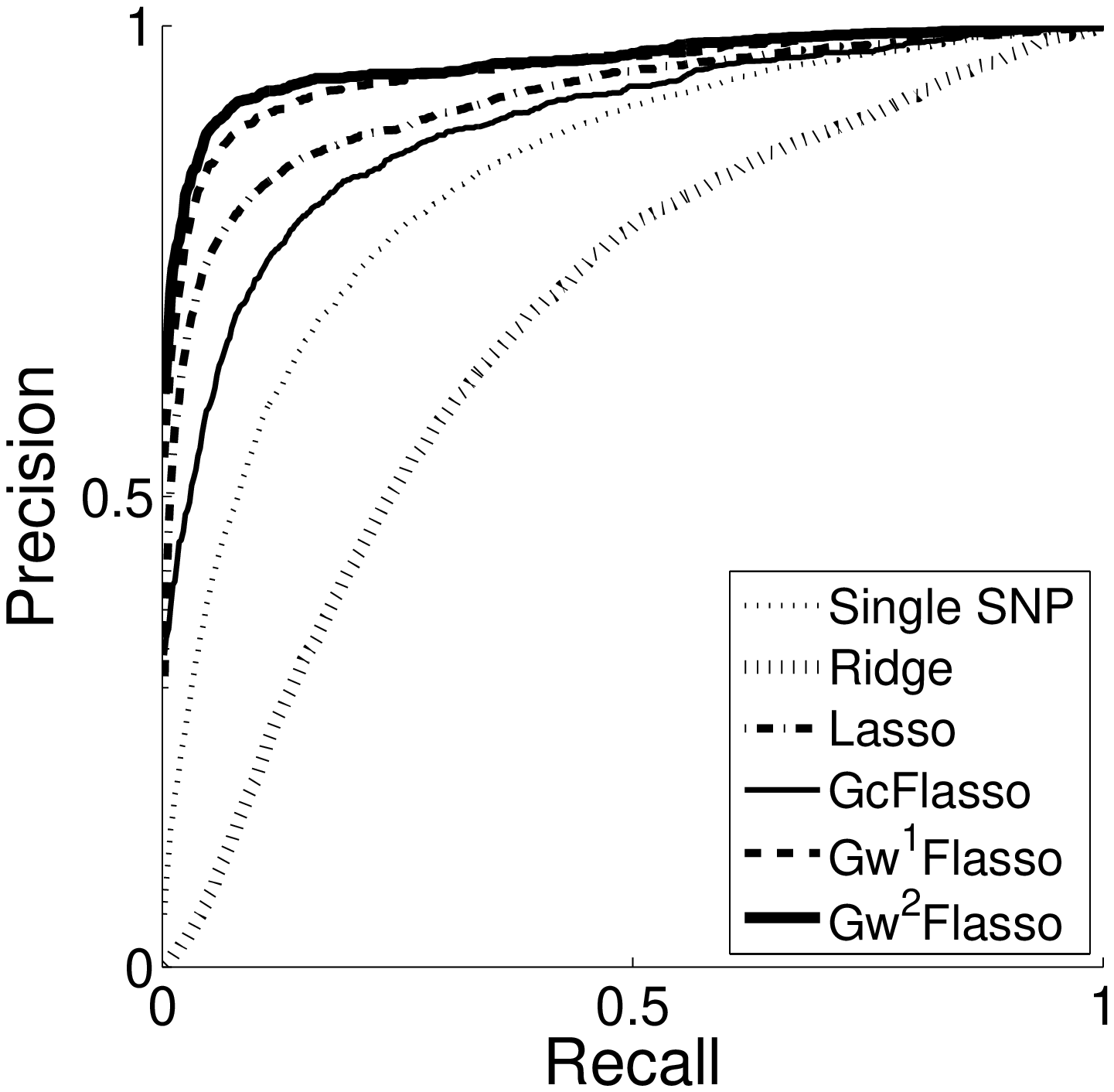} &
\includegraphics[scale = 0.27]{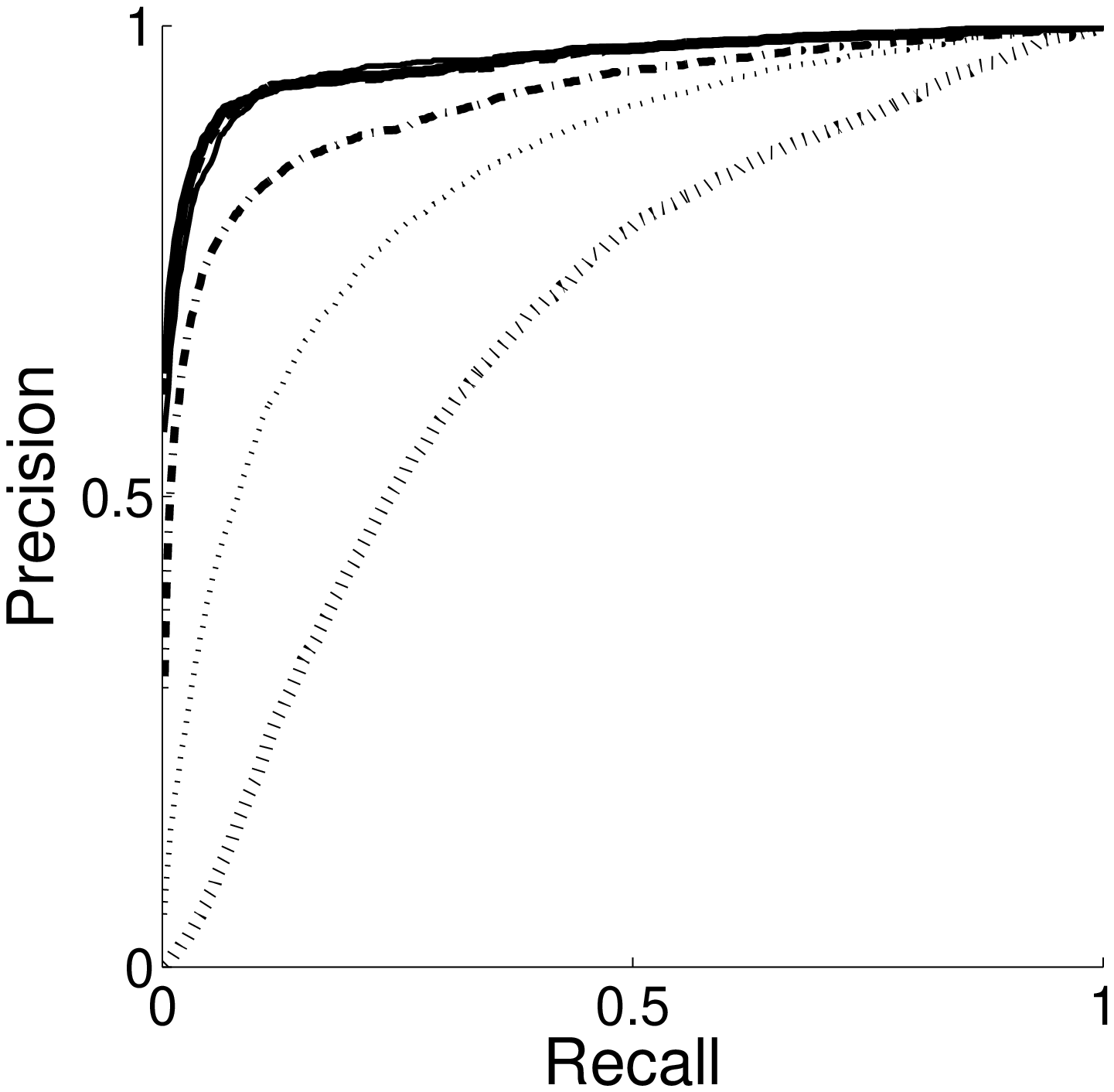} &
\includegraphics[scale = 0.27]{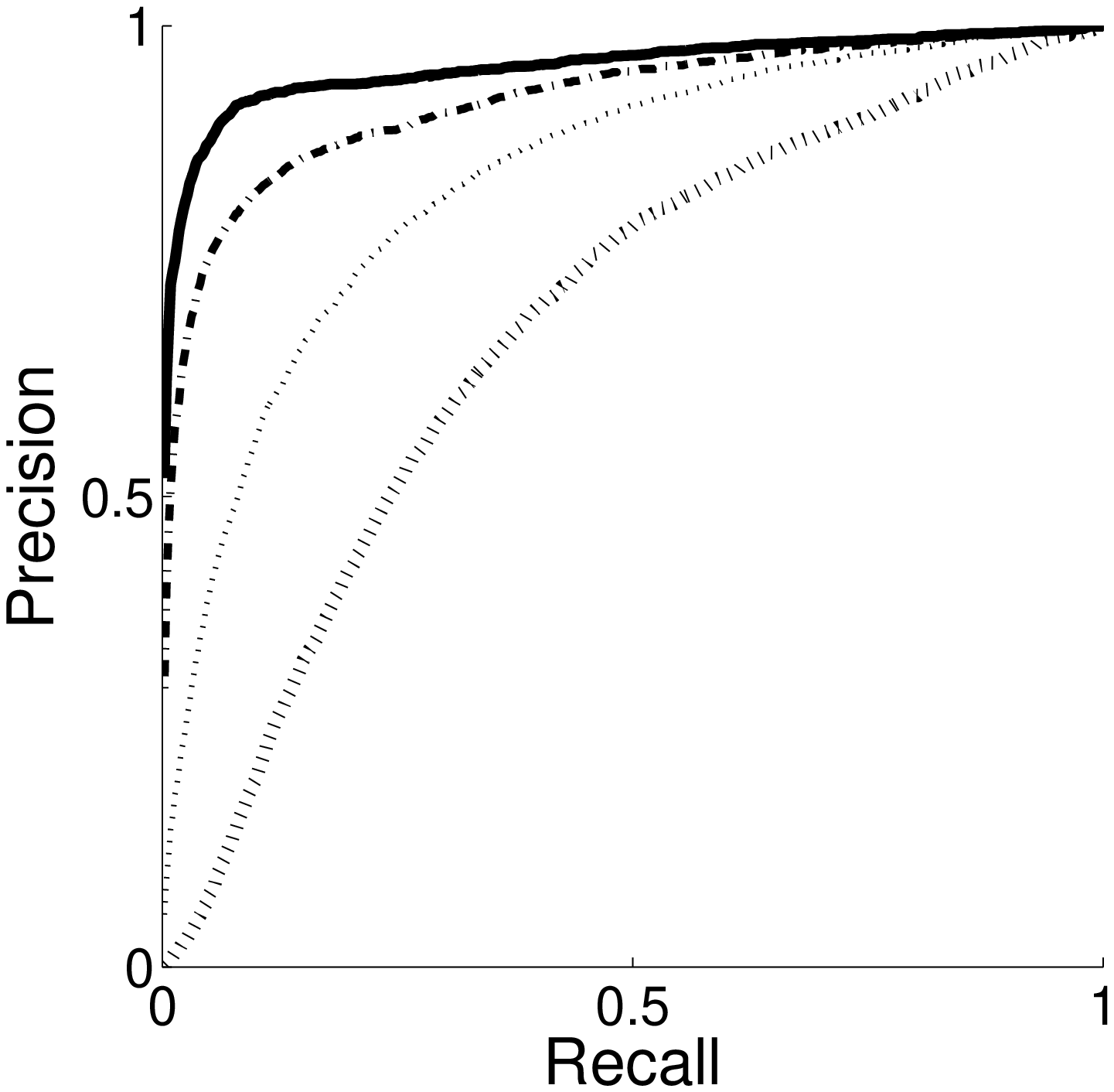} &
\includegraphics[scale = 0.27]{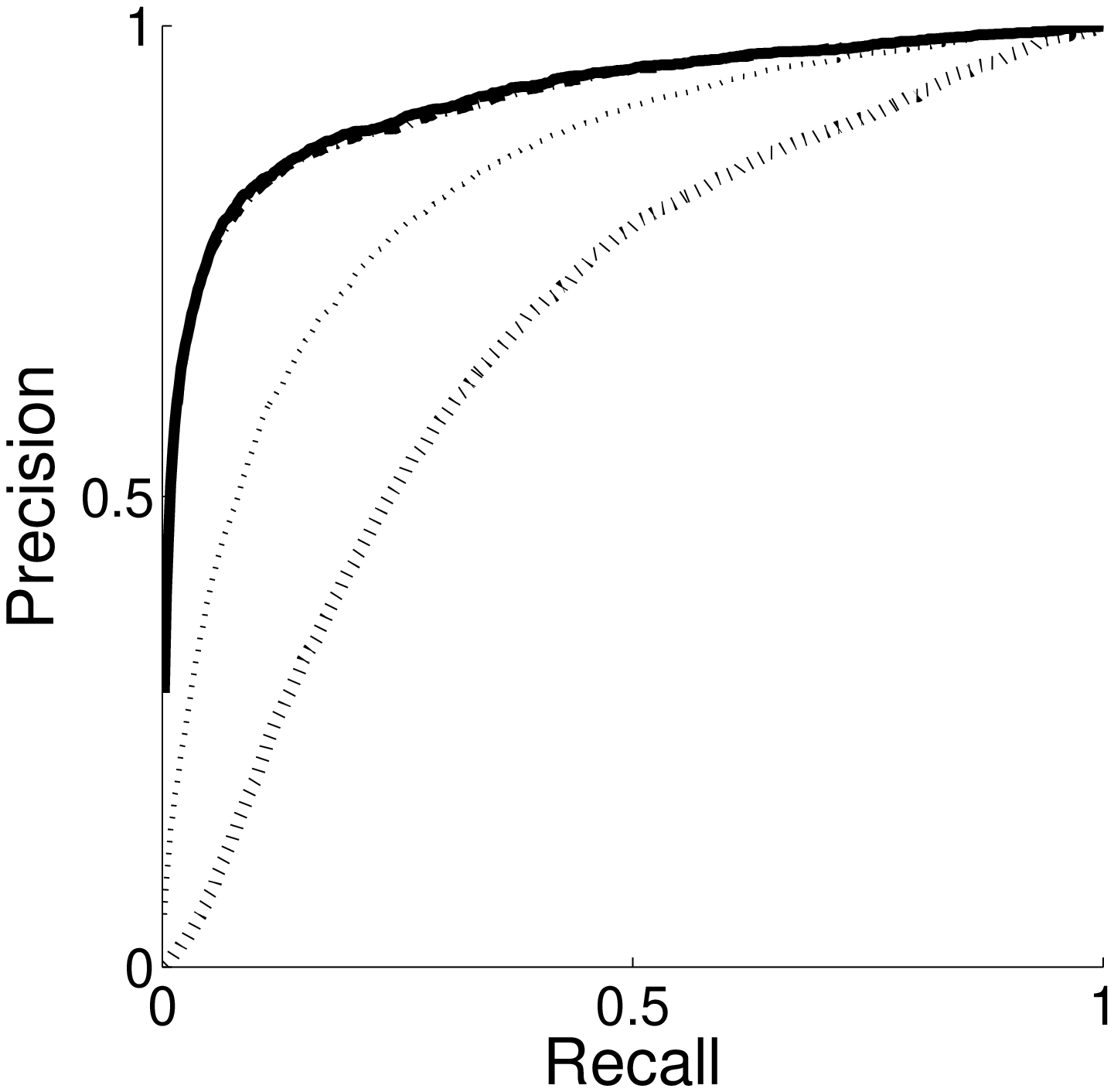} \\
A & B & C & D
\end{tabular}
\caption{
ROC curves for comparison of association analysis methods
with different values of threshold ($\rho$) for the phenotype correlation network.
A: $\rho$=0.1, B: $\rho$=0.3, C: $\rho$=0.5, and D: $\rho$=0.7.
The sample size is 100, and the effect size is 0.8.
} \label{fig:sim_th}
\end{figure}

In the results shown below, for each dataset of size $N$,
we fit the lasso and the graph-guided methods
using $(N-30)$ samples, and use the remaining 30 samples as a validation set
for determining the regularization parameters. Once we determine the regularization
parameters, we use the entire dataset of size $N$ to estimate the
final regression coefficients given the selected regularization parameters.

\begin{figure}[t!]\centering
\begin{tabular}{@{}c@{}c@{}c@{}c}
\includegraphics[scale = 0.23]{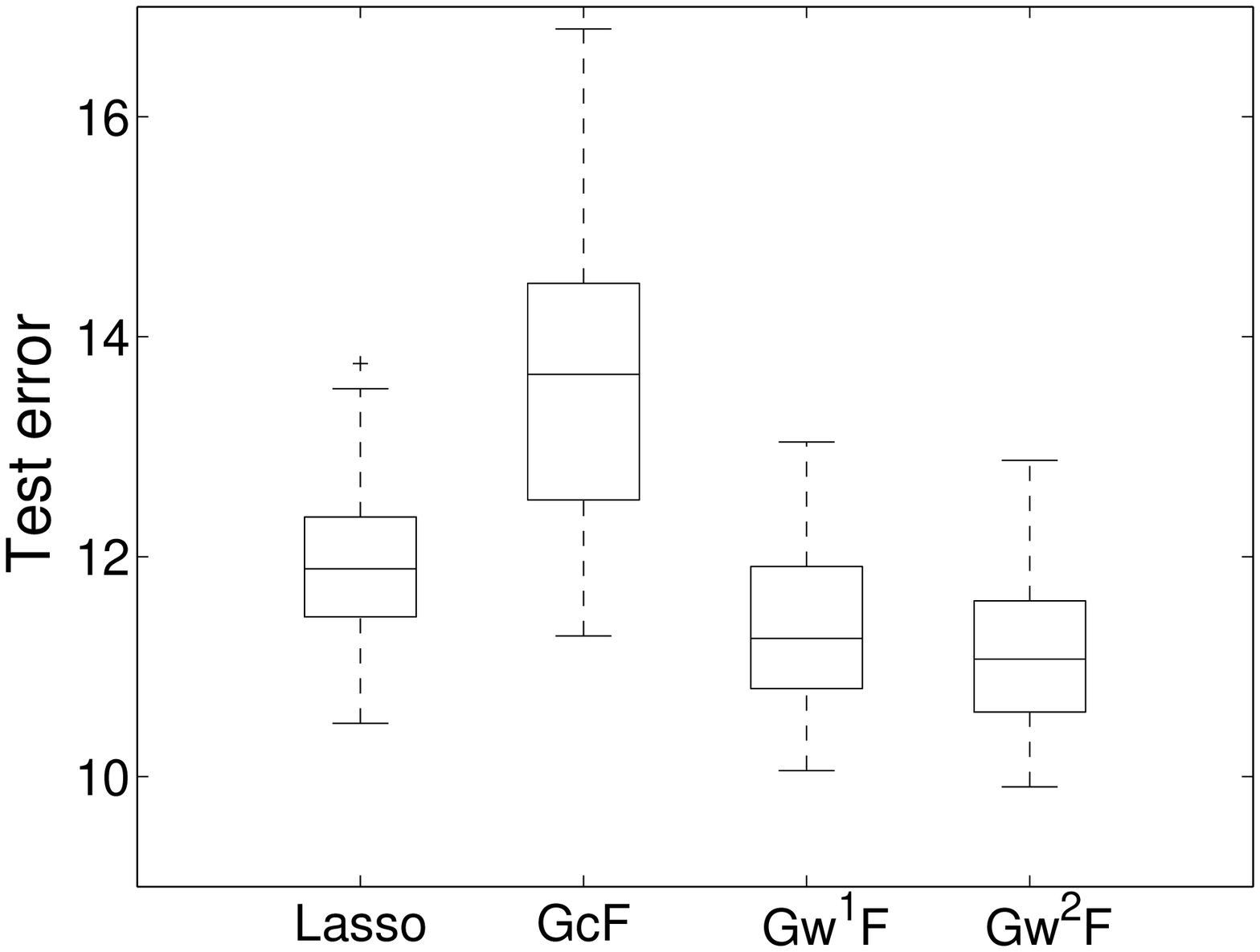} &
\includegraphics[scale = 0.23]{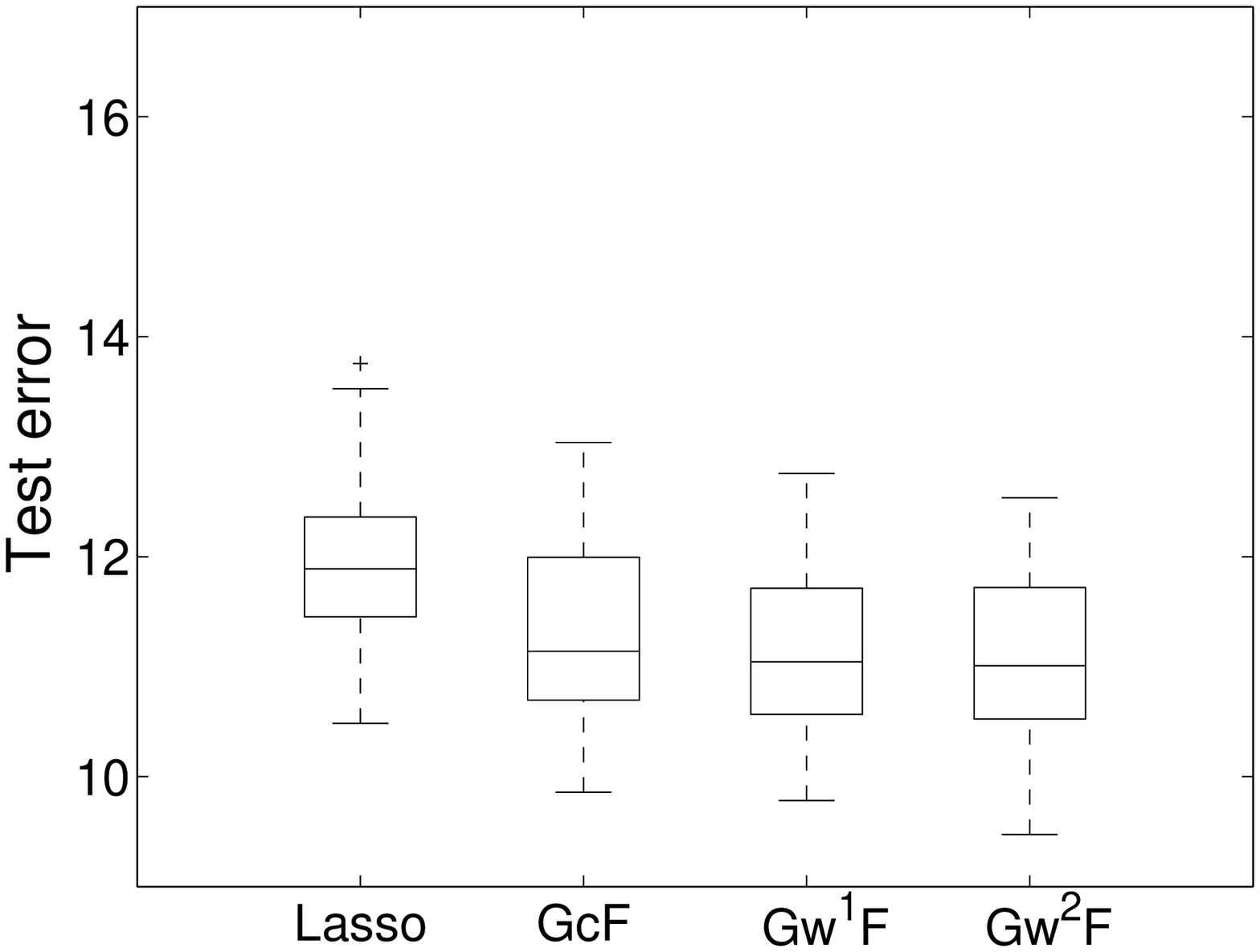} &
\includegraphics[scale = 0.23]{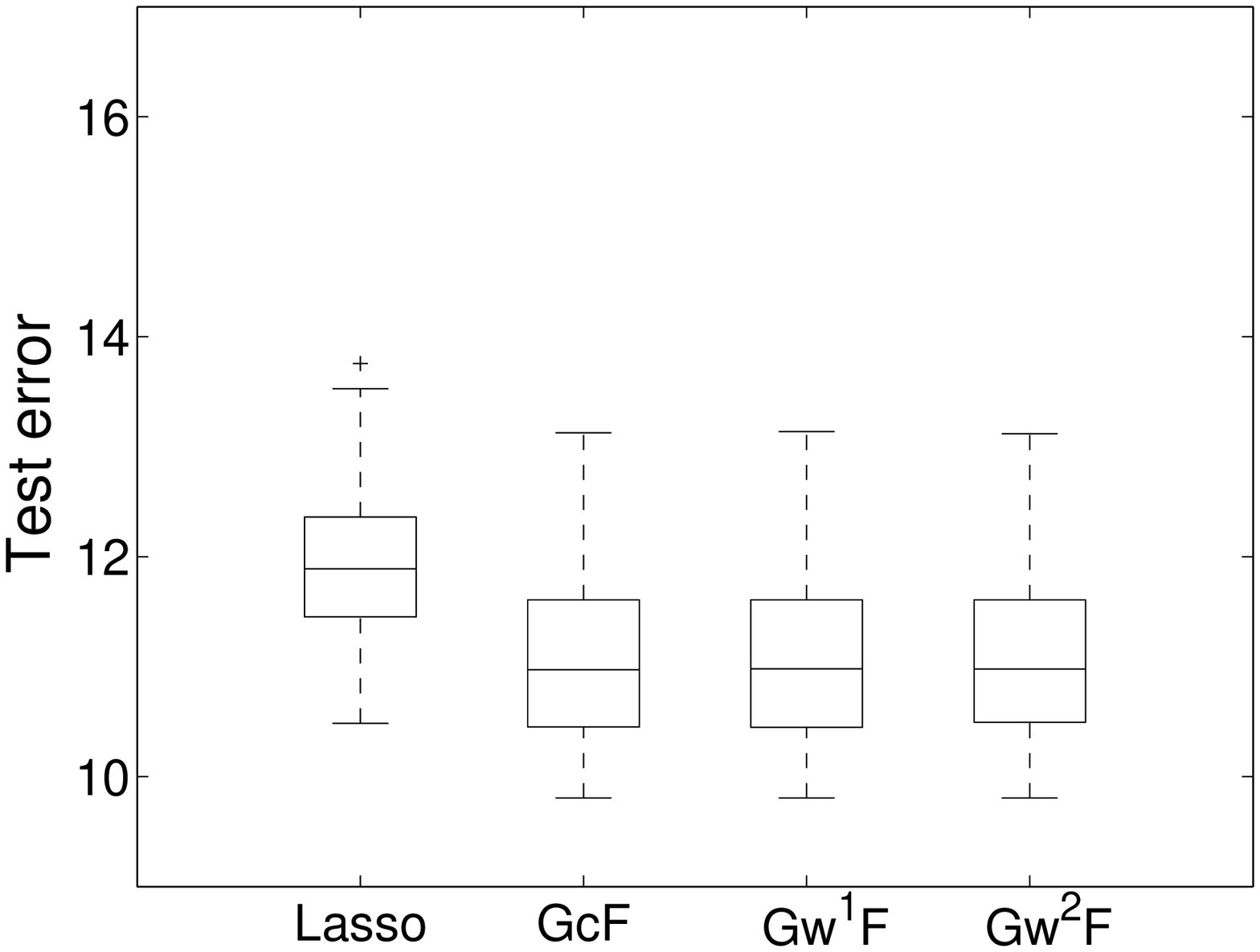} &
\includegraphics[scale = 0.23]{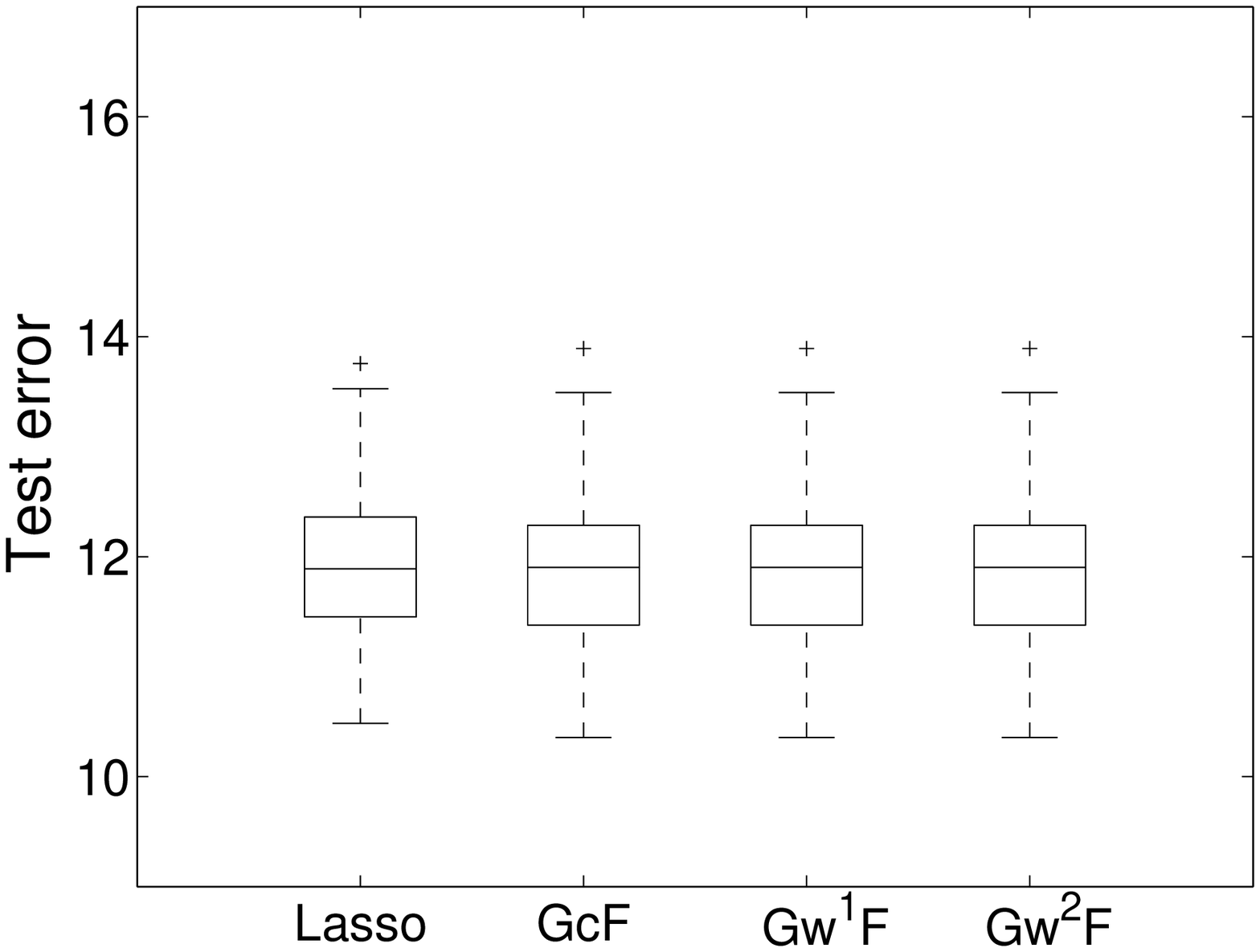} \\
A & B & C & D
\end{tabular}
\caption{ Comparison of association analysis methods
in terms of phenotype prediction error.
The threshold $\rho$ for the phenotype correlation network is
A: $\rho$=0.1, B: $\rho$=0.3, C: $\rho$=0.5, and D: $\rho$=0.7.
}
\label{fig:sim_ts_err}
\end{figure}

\begin{figure}[t!]\centering
\begin{tabular}{cccc}
\hspace{3pt} \includegraphics[scale = 0.20]{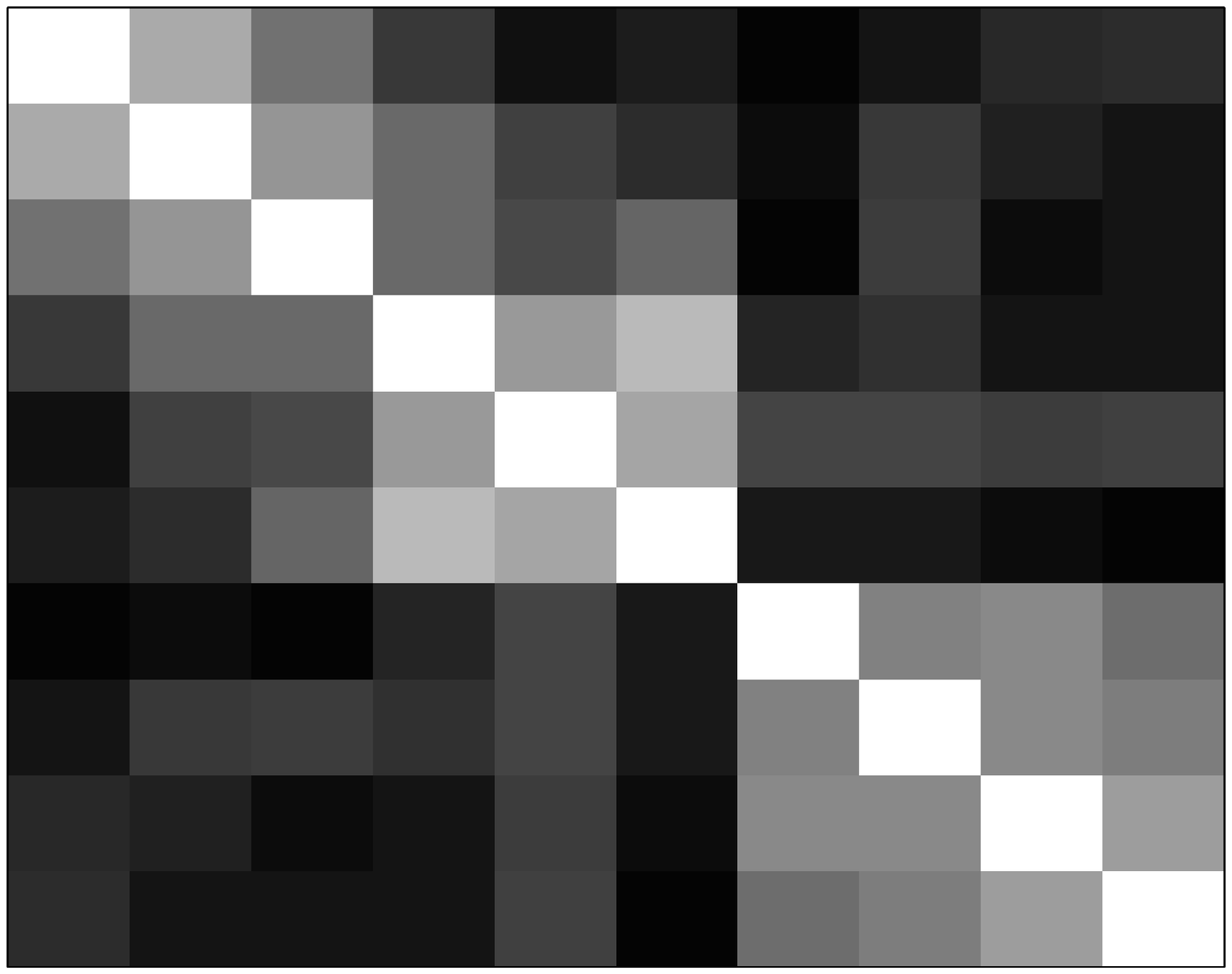} &
\hspace{2pt} \includegraphics[scale = 0.20]{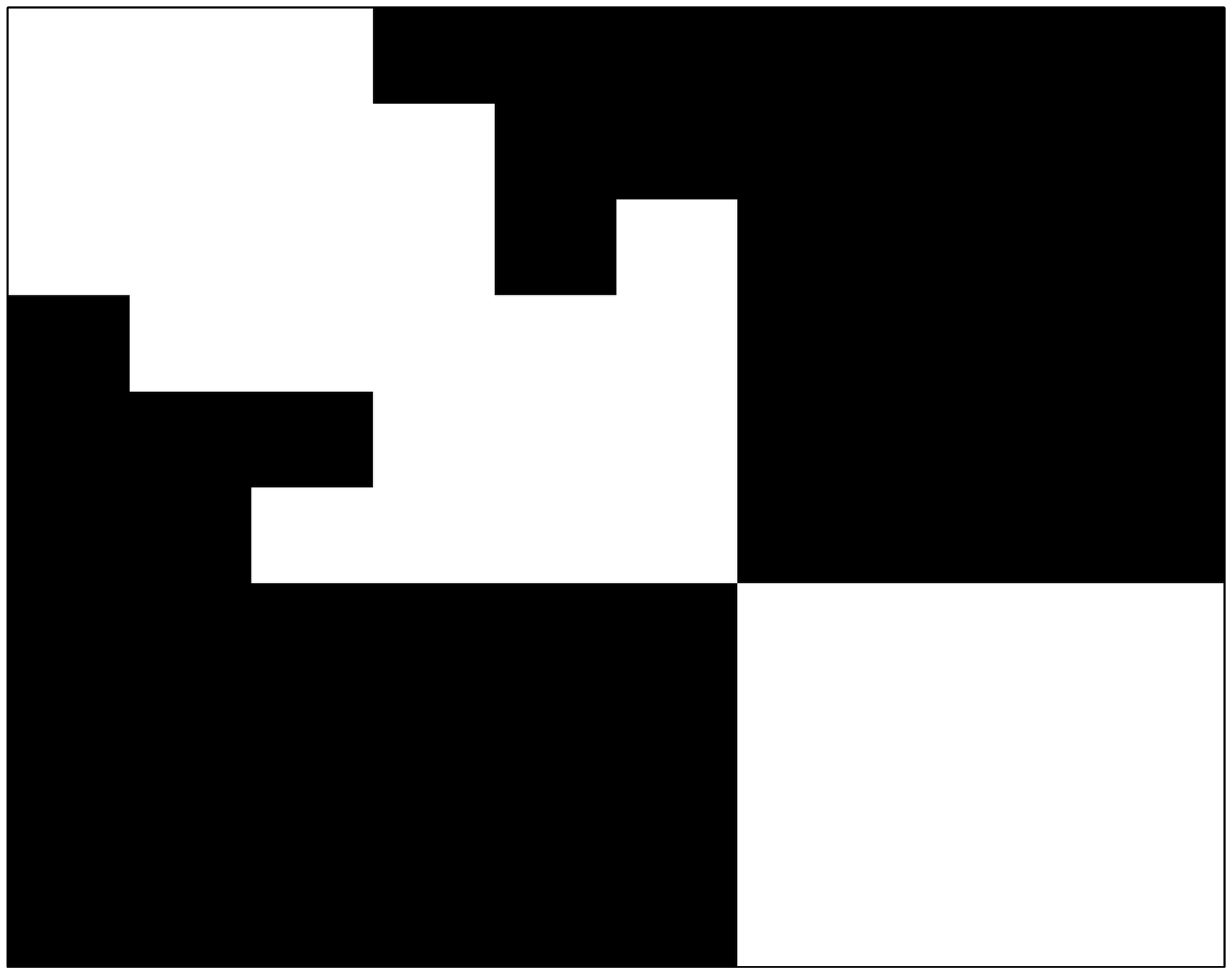} & &  \\
A & B & &
\vspace{5pt} \\
\includegraphics[scale = 0.20]{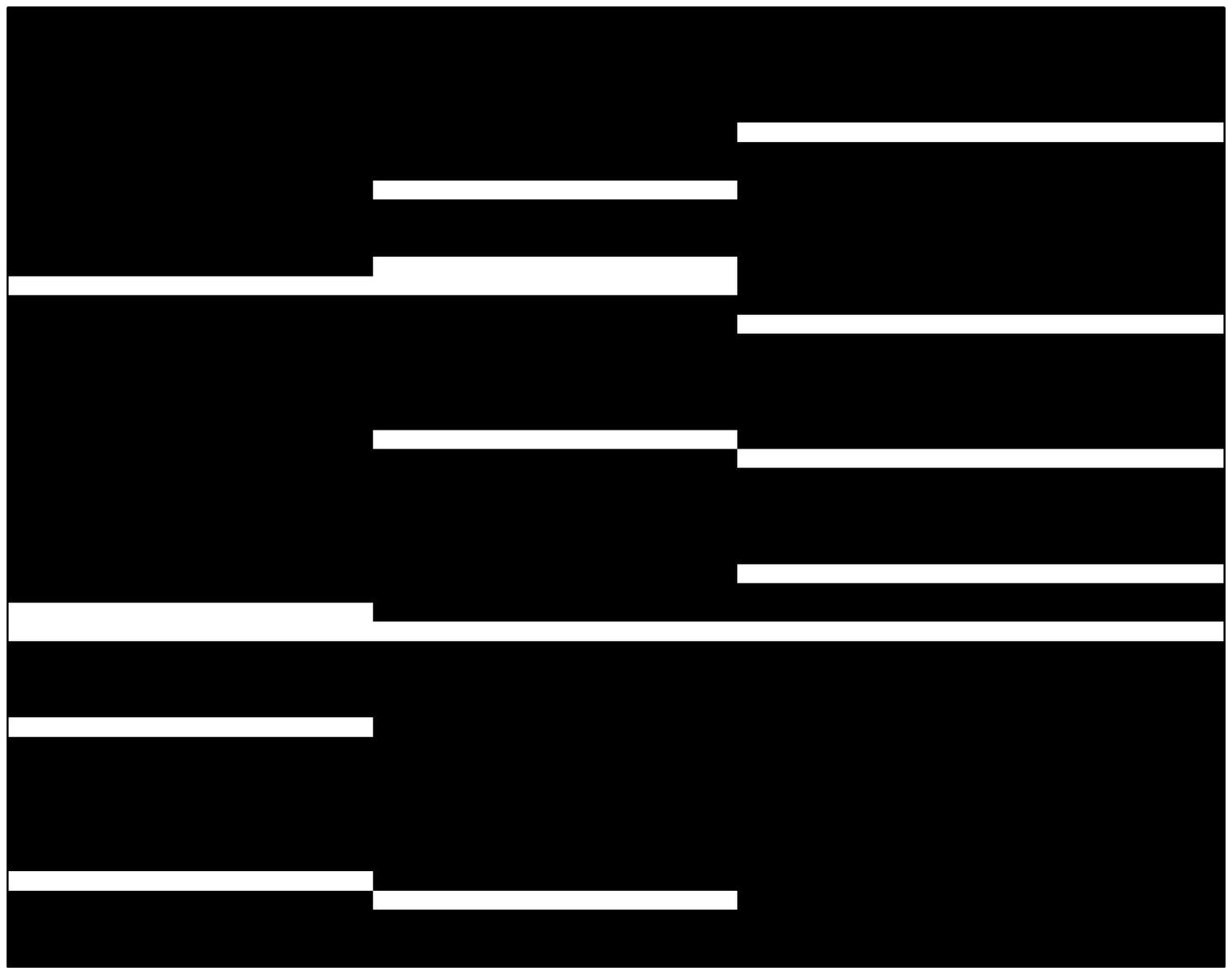} &
\includegraphics[scale = 0.20]{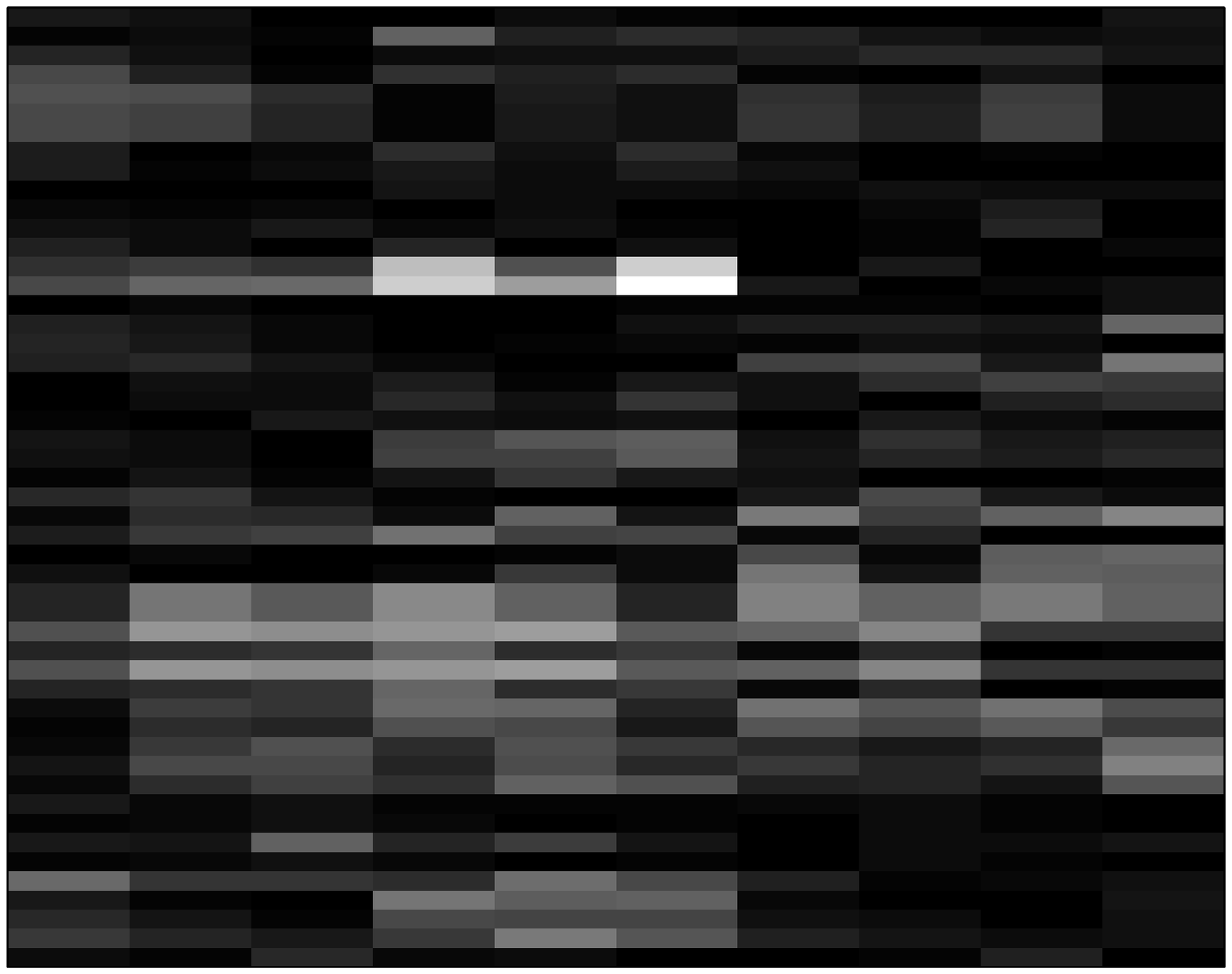} &
\includegraphics[scale = 0.20]{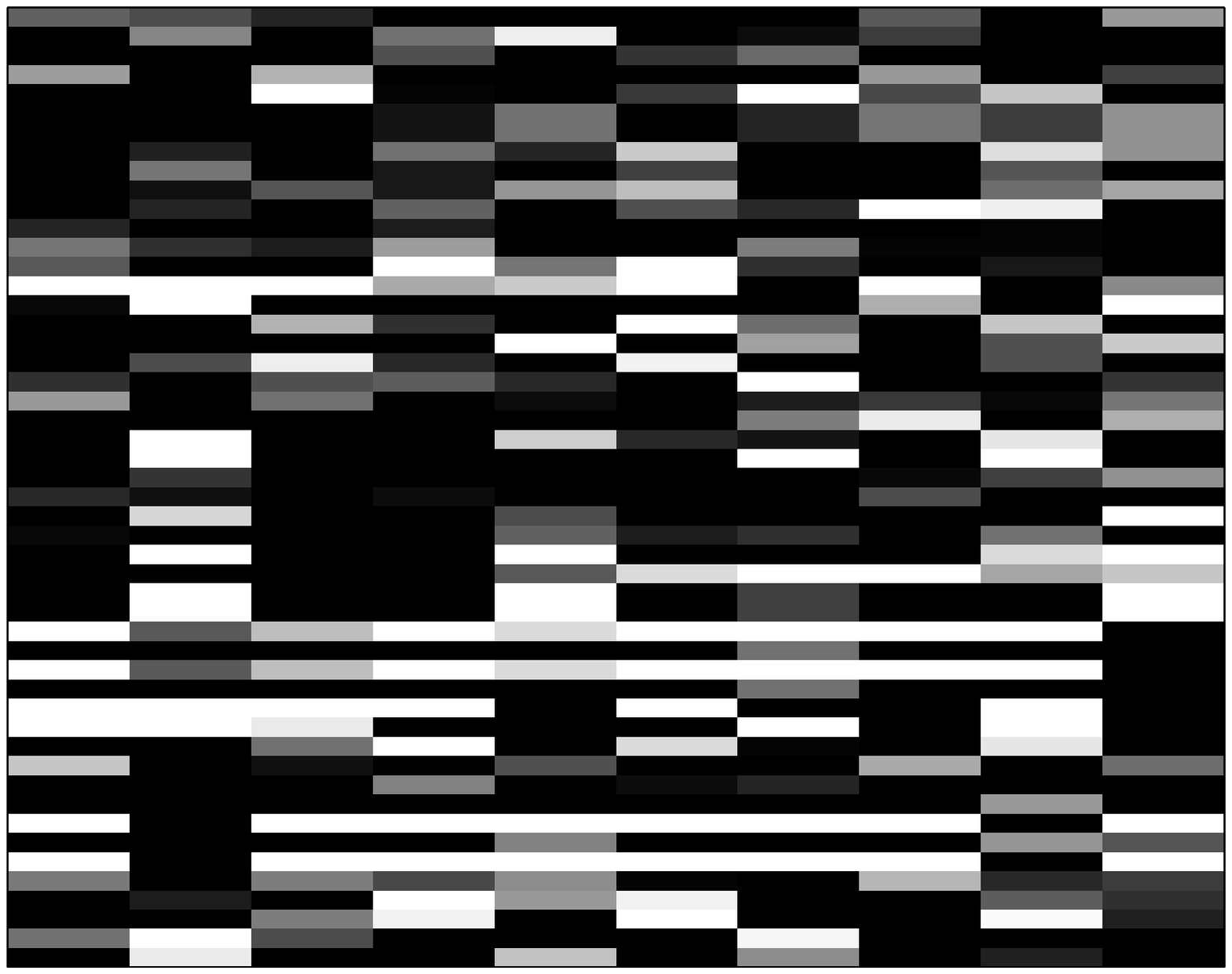} & \\
C & D & E &
\vspace{5pt} \\
\includegraphics[scale = 0.20]{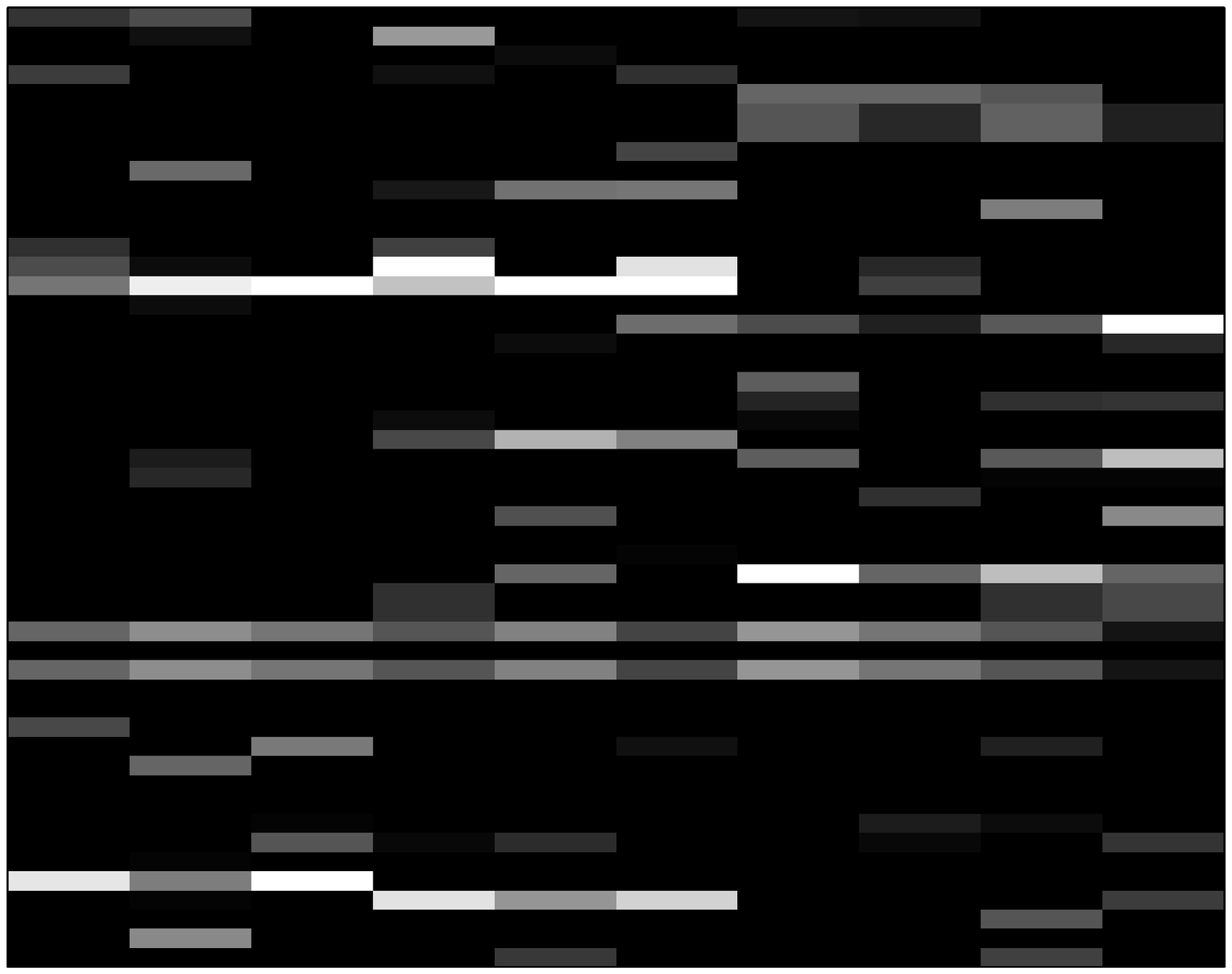} &
\includegraphics[scale = 0.20]{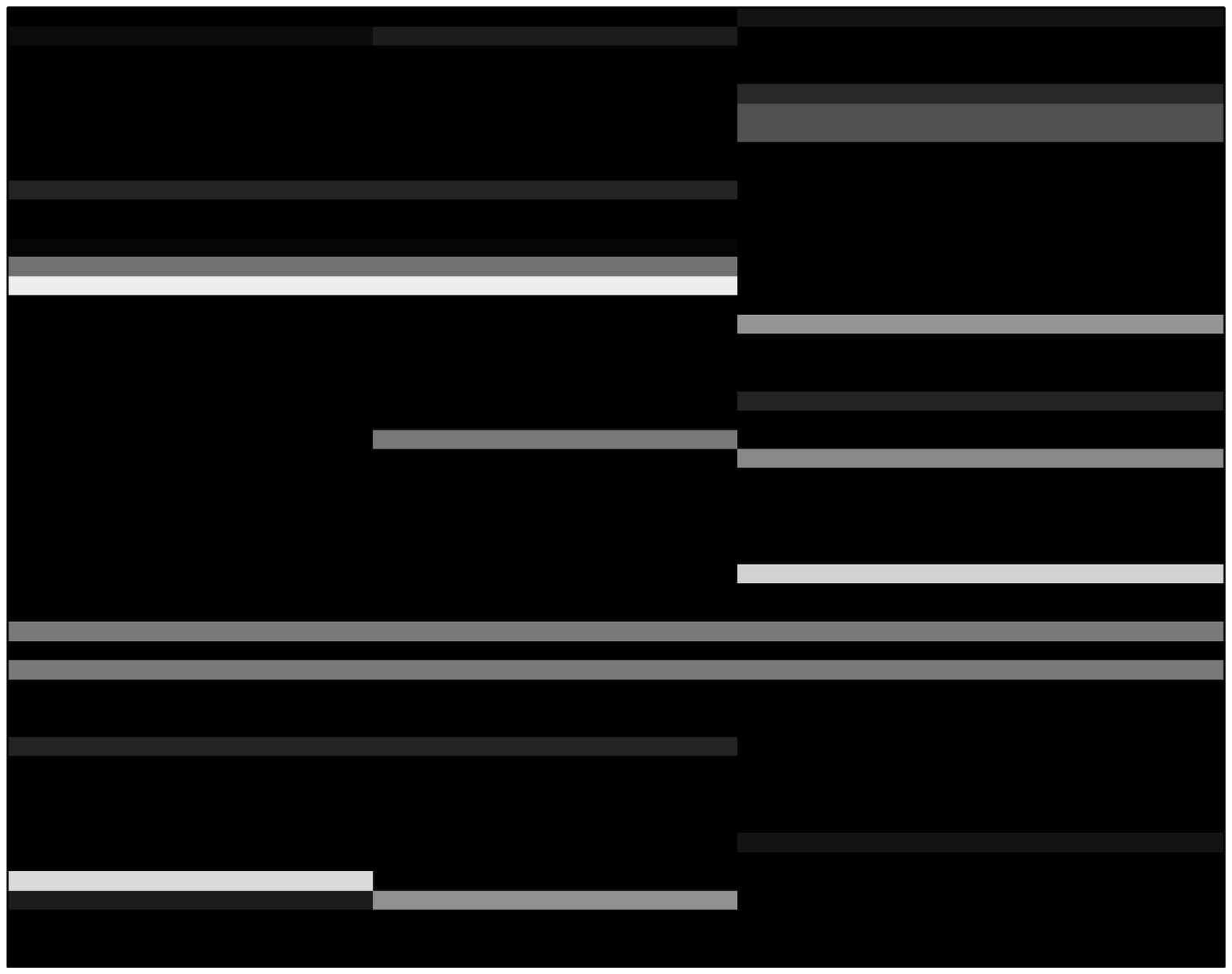} &
\includegraphics[scale = 0.20]{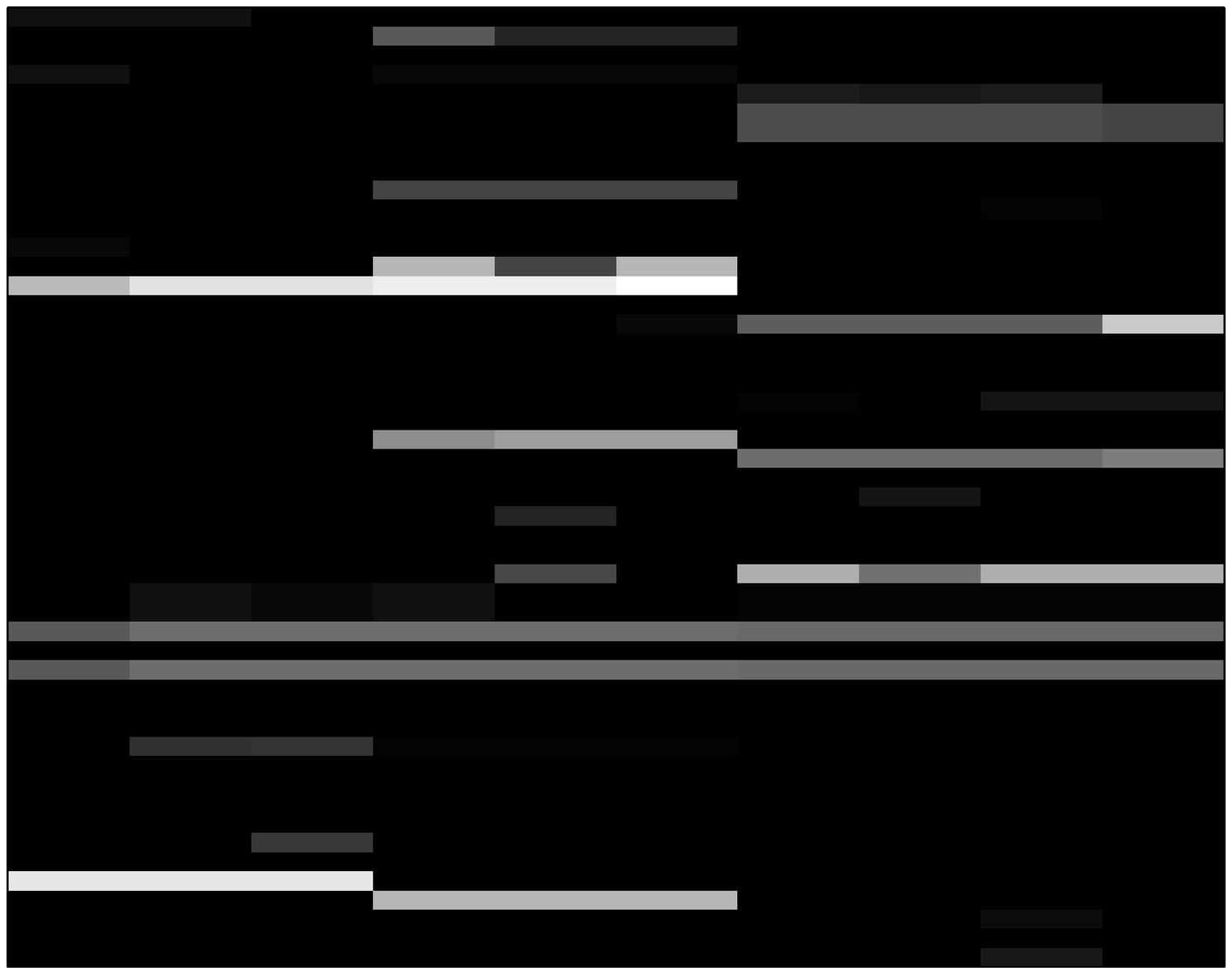} &
\includegraphics[scale = 0.20]{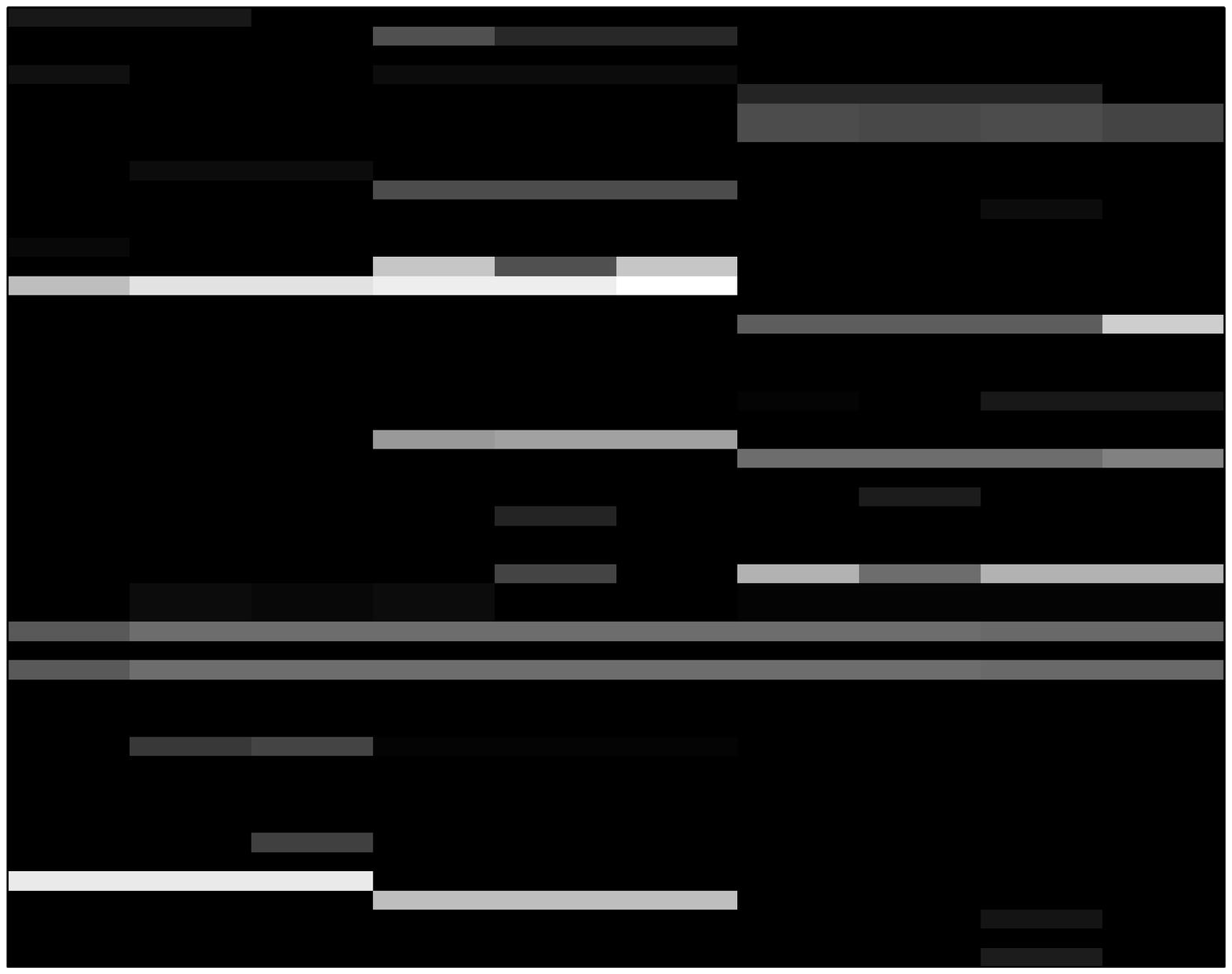} \\
F & G & H & I
\vspace{5pt} \\
\end{tabular}
\caption{Results of association analysis by different methods
based on a single simulated dataset. Effect size 0.8 and threshold $\rho=0.3$
for the phenotype correlation graph are used. Bright pixels indicate
large values. A: The correlation coefficient matrix of phenotypes,
B: the edges of the phenotype correlation graph obtained at threshold 0.3
are shown as white pixels, C: The true regression coefficients used
in simulation. Rows correspond to SNPs and columns to phenotypes.
D: -log($p$-value). Absolute values of the estimated regression coefficients
are shown for E: ridge regression, F: lasso, G: \Gc, H: \Gw, and I: \Gww.
 } \label{fig:sim_b_img}
\end{figure}

We apply the various association methods to datasets with varying sample
sizes, and show the ROC curves in Figure \ref{fig:sim_ss}.
We used the threshold $\rho$=0.3 to obtain the phenotype correlation
graph, and set the effect size to 0.5. The results confirm that
the lasso is an effective method for detecting true causal SNPs
and is affected less by the irrelevant SNPs, compared to the single-marker
analysis and ridge regression. When we use the weighted fusion penalty
in addition to the lasso penalty as in \Gc, \Gw, and \Gww,
the performance significantly improves over the lasso
across all of the samples sizes shown in Figure \ref{fig:sim_ss}.

\begin{figure}[t!]\centering
\begin{tabular}{cc}
\includegraphics[scale = 0.34]{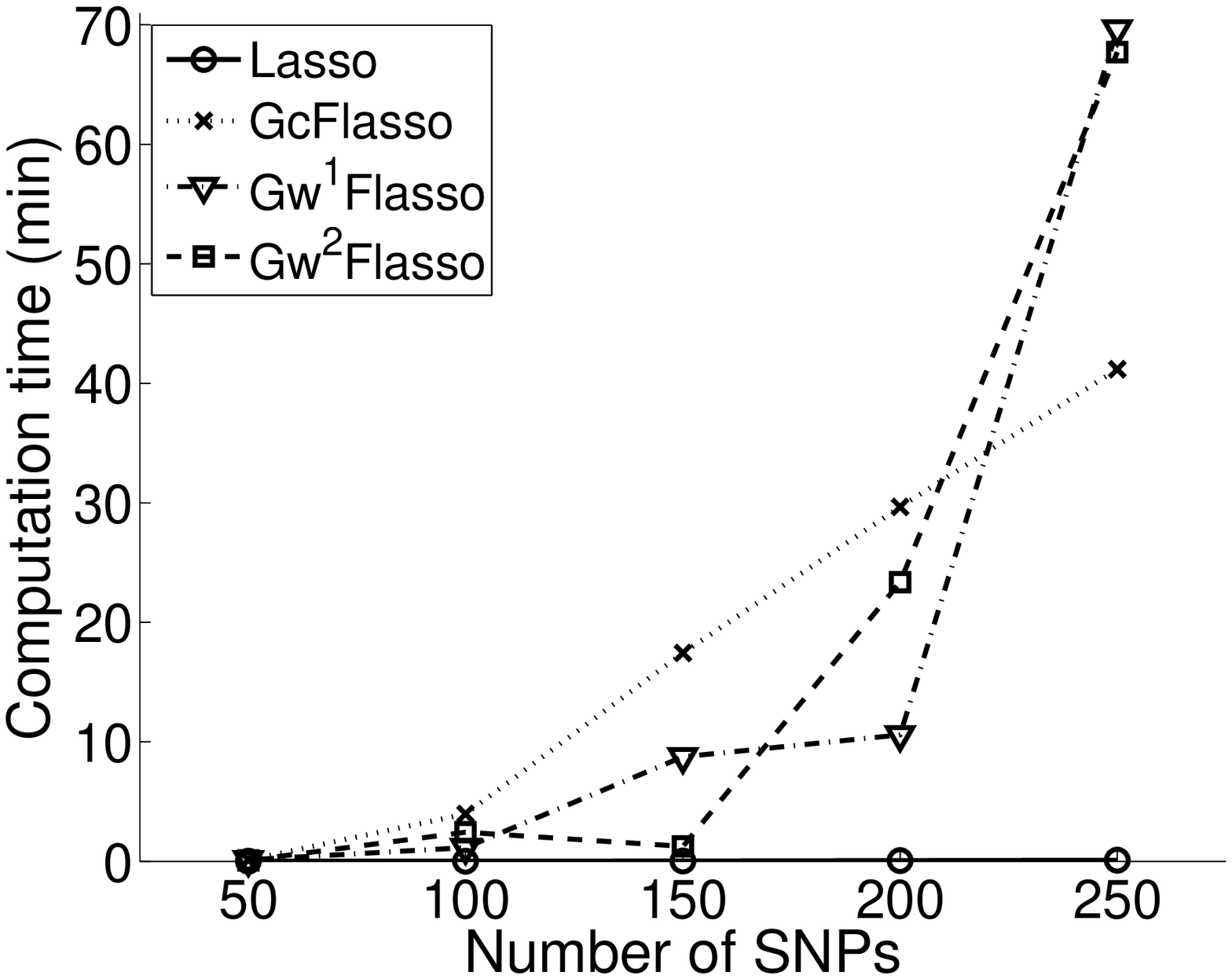} &
\includegraphics[scale = 0.34, bb=30 180 540 590]{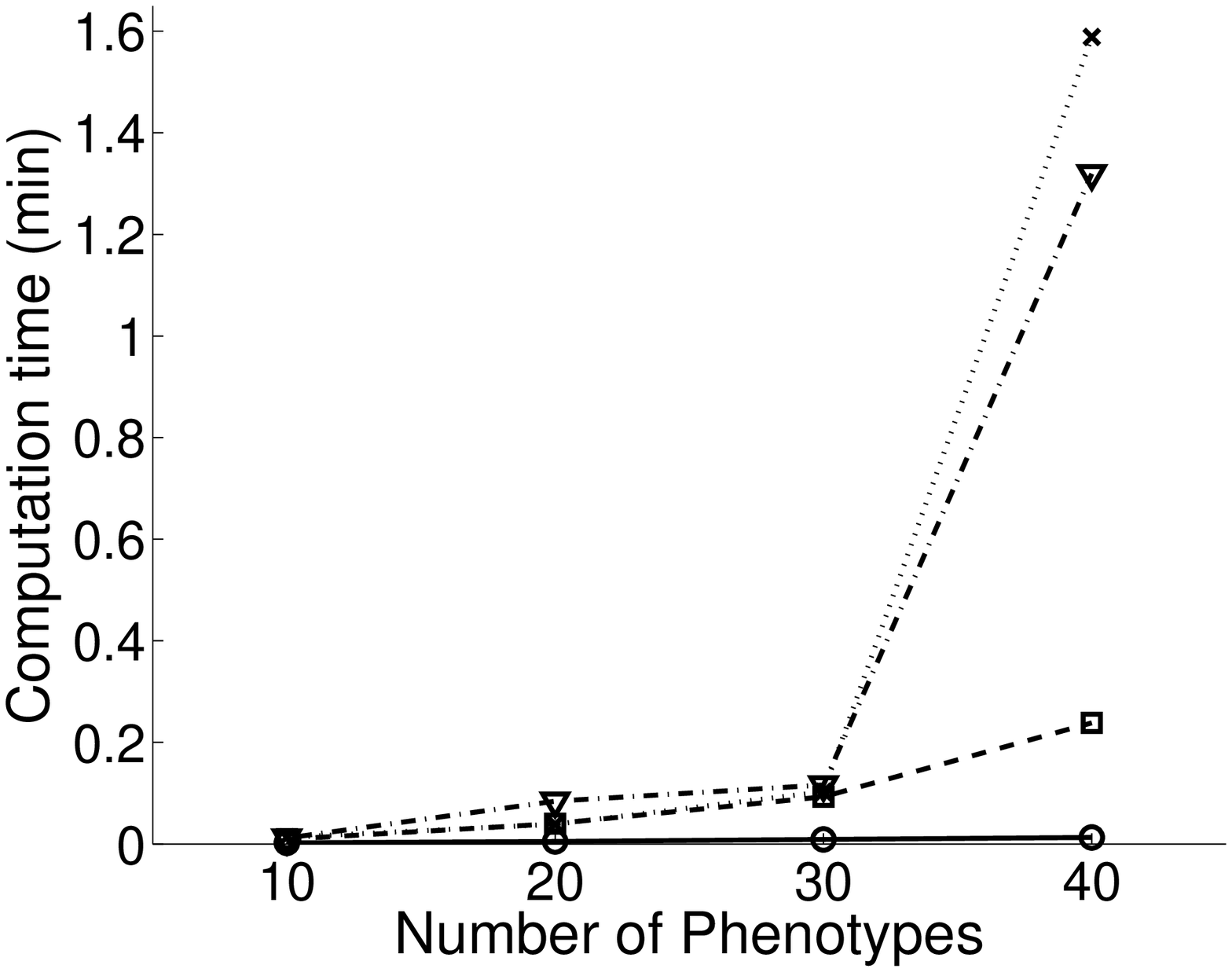} \\
A & B
\end{tabular}
\caption{ Comparison of the computation time for lasso, \Gc, \Gw, and \Gww.
A: Varying the number of SNPs with the number of phenotypes fixed at
10. The phenotype correlation graph at threshold $\rho=0.3$ with 31 edges is used.
B: Varying the number of phenotypes with the number of SNPs fixed at 50.
The phenotype networks are obtained using threshold $\rho=0.3$. The number
of edges in each phenotype network is 11, 34, 53, 88, and 142 for
the number of phenotypes 10, 20, 30, 40, and 50, respectively.
} \label{fig:sim_time}
\end{figure}

In order to see how the effect size affects the performance of the
methods for association analysis, we vary the effect size and
show the ROC curves in Figure \ref{fig:sim_b},
for the threshold $\rho$=0.1 of the phenotype
correlation network and sample size $N=100$. The \Gw~ and \Gww~
outperform all of the other methods across all of the effect sizes.
Because of the relatively low value of the threshold $\rho=0.1$,
the correlation phenotype contains many edges between a pair of phenotypes
that are only weakly correlated. Thus, the \Gc~ that does not distinguish
edges for strong correlation from those for weak correlation does not
show a consistent performance across different effect size, performing
better than the lasso at effect size 0.3 but worse than the lasso at effect size 1.0.
The \Gw~ and \Gww~ have the flexibility to handle different strengths
of correlation in the graph, and consistently outperforms \Gc~
as well as the methods
that do not consider the structural information in the phenotypes.

In order to examine the effect of the threshold $\rho$
for the phenotype correlation graph on the performance of our methods,
we evaluate the GFlasso methods with $\rho$ at 0.1, 0.3, 0.5, and 0.7, and show the
ROC curves in Figure \ref{fig:sim_th}. We include the
ROC curves for the single-marker analysis, the ridge regression,
and the lasso that do not use the thresholded phenotype correlation
graph in each panel of Figure \ref{fig:sim_th} repeatedly for the ease of comparison.
We use the sample size $N=100$ and the effect size 0.8. Regardless of
the threshold $\rho$, the \Gw~ and \Gww~ outperform all of the other methods
or perform at least as well as the lasso.
As we have seen in Figure \ref{fig:sim_b}, the \Gc~ does not have
the flexibility of accommodating edges of varying correlation strength
in the phenotype correlation graph, and this negatively affects the
performance of the \Gc~ at the low threshold $\rho=0.1$ in Figure \ref{fig:sim_b}A.
As we increase the threshold $\rho$ in Figures \ref{fig:sim_b}B and C,
the phenotype correlation graph include only those edges with
significant correlations. Thus, the performance of \Gc~ approaches that of
\Gw~ and \Gww, and the curves of the three methods in the GFlasso family
almost entirely overlap. When the treshold is relatively high at $\rho=0.7$,
the number of edges in the graph is close to 0, effectively removing
the fusion penalty. As a result, the performance of the graph-guided methods
becomes close to the lasso. Overall, taking into account the correlation
structure in phenotypes improves the detection rate of true causal SNPs.
Once the phenotype correlation graph includes the edges that capture
strong correlations, including more edges by further lowering the threshold
$\rho$ does not significantly affect the performance of \Gw~ and \Gww.
The same tendency is shown in the prediction
errors in Figure \ref{fig:sim_ts_err}.

We show an example of a simulated dataset and the estimated association strength
in Figure \ref{fig:sim_b_img},  using the sample size $N=100$ and effect size 0.8.
Although the lasso is more successful in setting the regression coefficients
of irrelevant SNPs to zero than the ridge regression, it still finds many
SNPs as having a non-zero association strength.
The \Gc, \Gw, and \Gww~ remove most of those spurious SNPs, and shows a
clear block-structure in the estimated regression coefficients, with each
causal SNP spanning subgroups of of correlated phenotypes. Since
\Gc~uses only the information on the presence or absence of edges,
when edges of weak correlation connect nodes across two true subgraphs,
the \Gc~ is unable to ignore the weak edges, and fuses the effect of SNPs
on the phenotypes across those two subgraphs. This undesirable
property of \Gc~ disappears when we incorporate the edge weights in
\Gw~ and \Gww.

We show the computation time for solving a single optimization problem
for the lasso, \Gc, and \Gwa~ in Figure \ref{fig:sim_time} for varying number of
SNPs and phenotypes.

\subsection{Case Study Using Asthma Dataset}

\begin{figure}[t!]
\centering
\begin{tabular}{@{ }c@{ }c@{ }c}
\includegraphics[scale = 0.38]{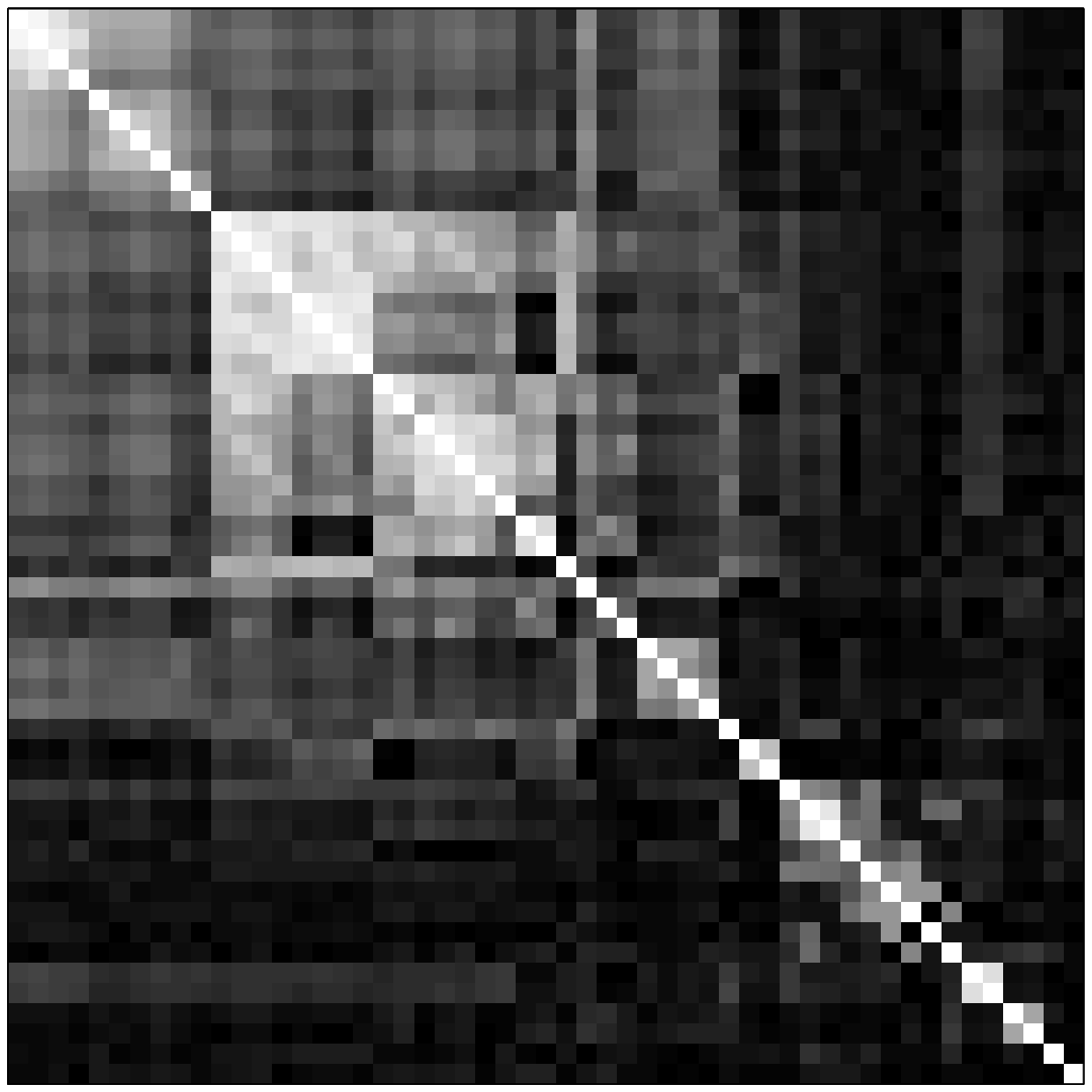} &
\includegraphics[scale = 0.38]{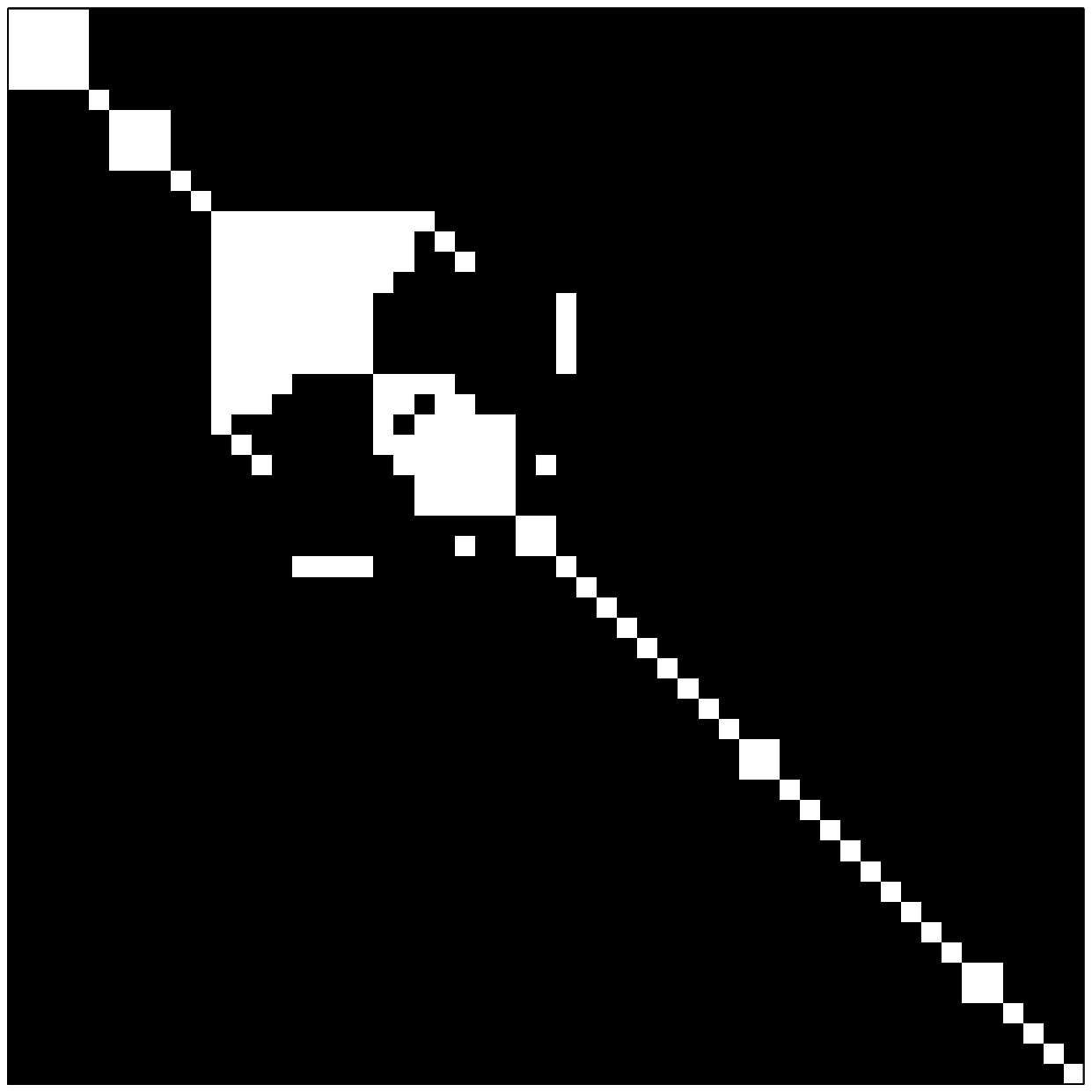} & \\
A & B &
\vspace{3pt}
\\
\includegraphics[scale = 0.38, angle = 90]{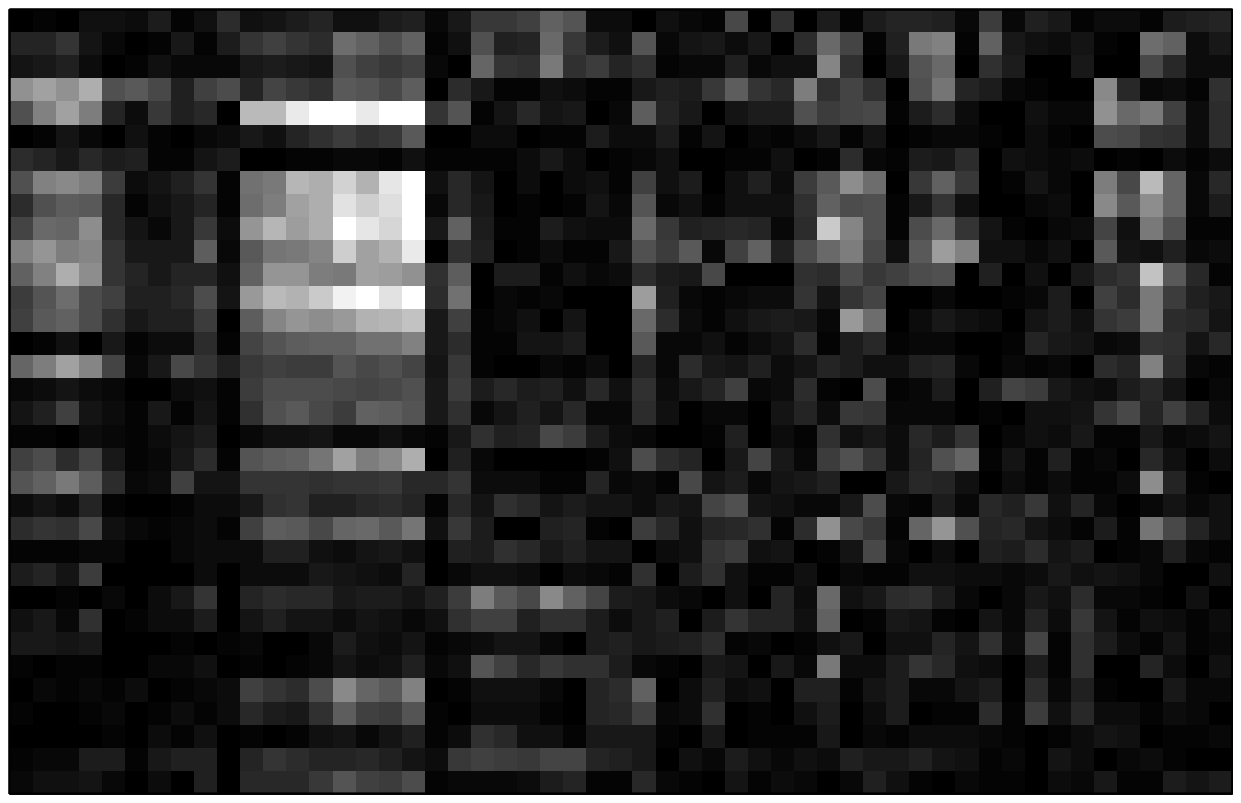} &
\includegraphics[scale = 0.38, angle = 90]{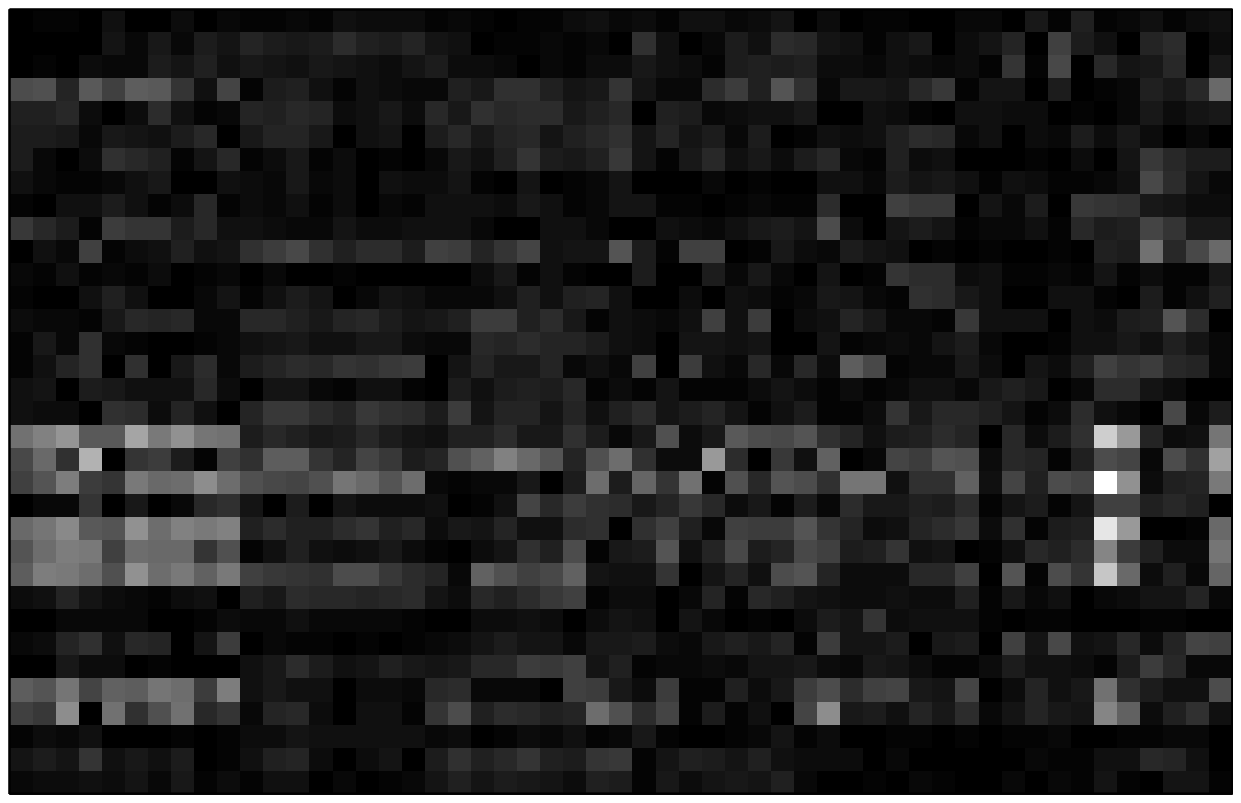} &
\includegraphics[scale = 0.38, angle = 90]{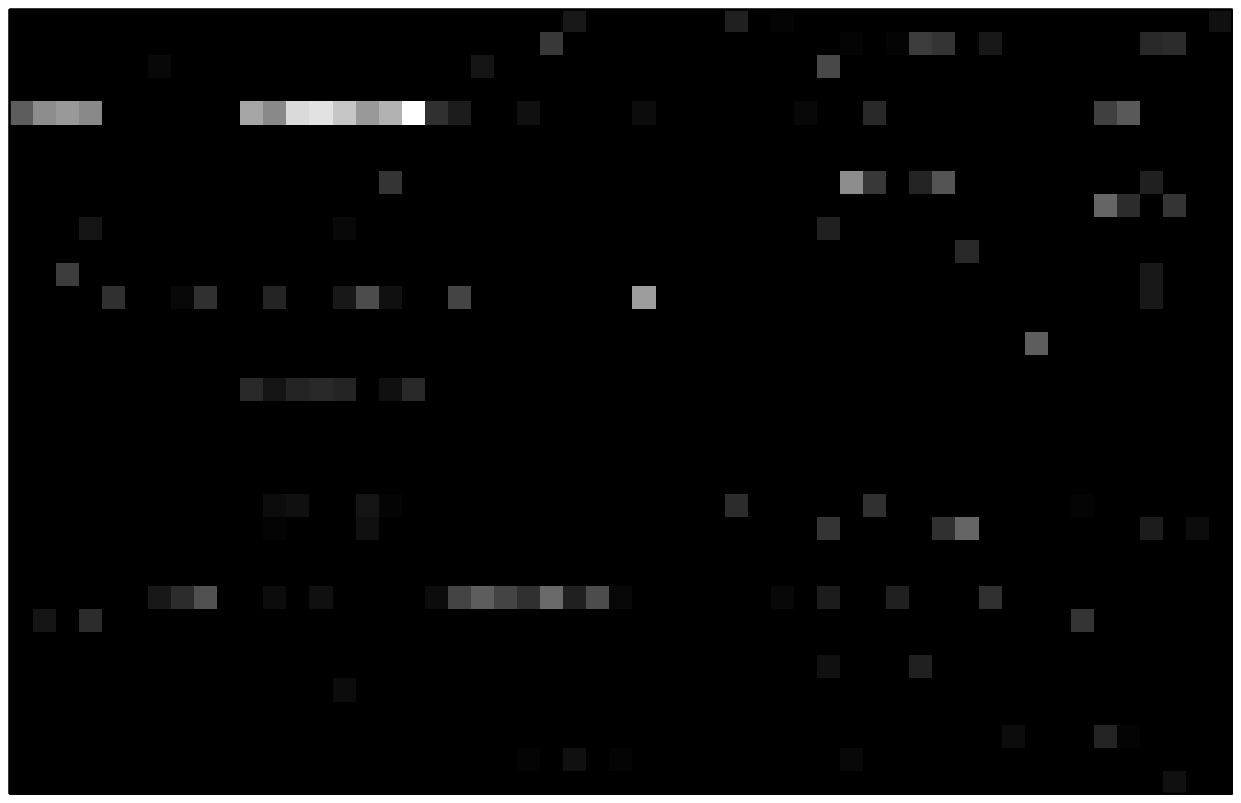}  \\
C & D & E
\vspace{3pt} \\
\includegraphics[scale = 0.38, angle = 90]{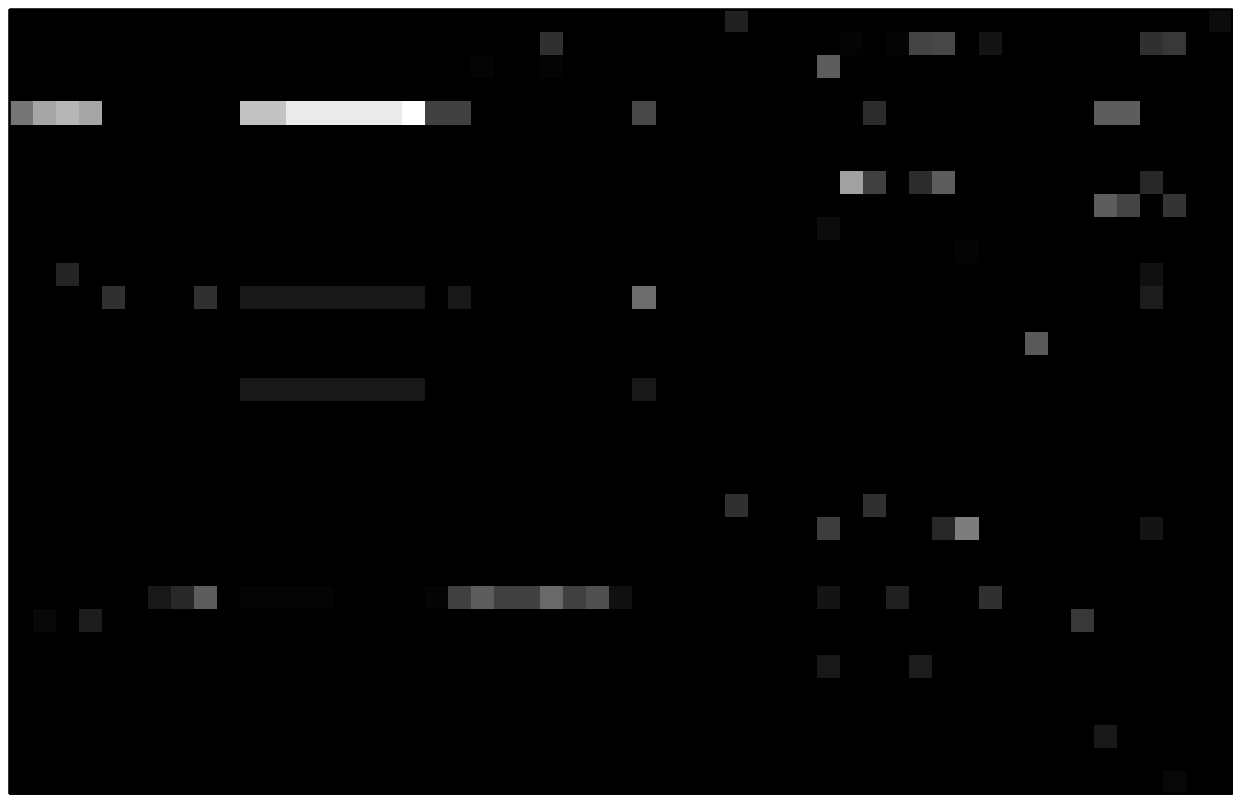} &
\includegraphics[scale = 0.38, angle = 90]{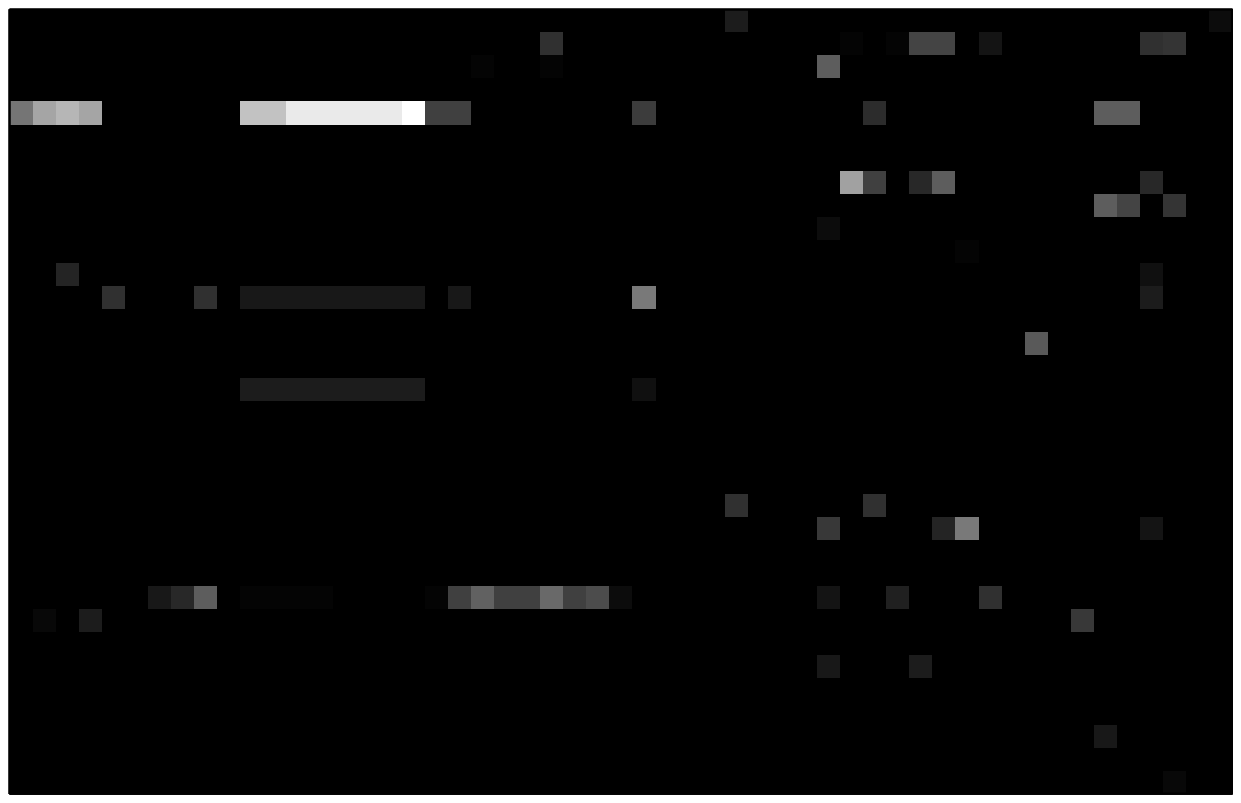} &
\includegraphics[scale = 0.38, angle = 90]{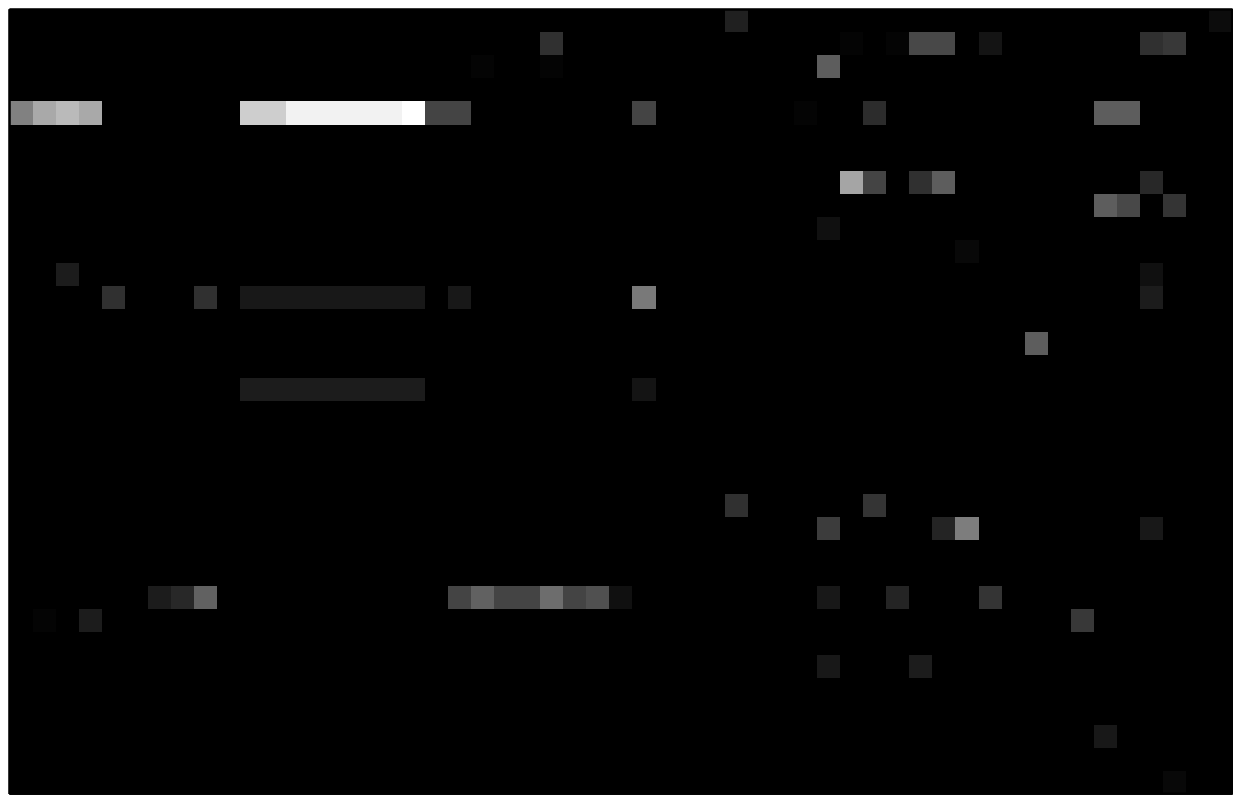} \\
F & G & H
\end{tabular}
\caption{
Results for the association analysis of the asthma dataset. A: Phenotype correlation matrix.
B: Phenotype correlation matrix thresholded at $\rho=0.7$.
C: -$\log$($p$-value) from single-marker
statistical tests using a single-phenotype analysis.
Estimated $\bm{\beta}_k$'s for D: ridge regression, E: lasso, 
F: \Gc, G: \Gw, and H: \Gww.
} \label{fig:sarp}
\end{figure}

Figure \ref{fig:sarp}A shows the
correlation matrix of the phenotypes after reordering the phenotypes
using the agglomerative hierarchical clustering algorithm so that
highly correlated phenotypes are clustered with a block structure
along the diagonal.
Using threshold $\rho=0.7$, we obtain a phenotype correlation graph
as shown in Figure \ref{fig:sarp}B, where the white pixel at
position $(i,j)$ indicates that the $i$-th and $j$-th phenotypes are
connected with an edge in the graph. The graph shows several blocks
of white pixels representing densely connected subgraphs. We show 
the full graph in Figure \ref{fig:sarp_g}.
We present results for the single-marker regression analysis,
the ridge regression, the lasso, \Gc, \Gw, and \Gww~ in Figure
\ref{fig:sarp}C-H, respectively, where the rows represent phenotypes,
and the columns correspond
to genotypes, with bright pixels indicating high strength of association.
The phenotypes in rows are rearranged according to the ordering given by the
agglomerative hierarchical clustering so that each row
in Figures \ref{fig:sarp}C-H is aligned with the phenotypes in
the correlation matrix in Figure \ref{fig:sarp}A.
In the fusion penalty in our proposed methods,
we use the edges in Figure \ref{fig:sarp}B obtained at threshold $\rho=0.7$.
The graph obtained at threshold $\rho=0.7$ seems to capture the previously
known dependencies among the clinical traits such as subnetworks 
corresponding to lung physiology and quality of life.
We select the regularization parameters
in the lasso, \Gc, \Gw, and \Gww~ using a five-fold cross validation.

As shown in Figures \ref{fig:sarp}C and E, both the single-marker
regression analysis and the lasso find a SNP near the top row, known as Q551R,
as significantly associated with a block of correlated phenotypes.
This subset of traits corresponds to the bottom subnetwork
(consisting of baselineFEV1, PreFEFPred, AvgNO, BMI, PostbroPred,
BaseFEVPer, PredrugFEV1P, MaxFEV1P, FEV1Diff, and PostFEF)
that resides within the large subnetwork on the left-hand side of Figure \ref{fig:sarp_g},
and represents traits related to lung physiology.
This Q551R SNP has been previously found associated with severe
asthma and its traits for lung physiology~\cite{Wenz:2007}, and our results
confirm this previous finding. In addition, the results from
the single-marker analysis in Figure \ref{fig:sarp}C
show that on the downstream of this SNP, there is
a set of adjacent SNPs that appears to be in linkage disequilibrium
with this SNP and at the same time has generally a high level
of association with the same subset of phenotypes.
On the other hand,
the lasso in Figure \ref{fig:sarp}E sets most of the regression
coefficients for this block of SNPs in linkage disequilibrium with Q551R to zero, identifying
a single SNP as significant. This confirms that the lasso is
an effective method for finding sparse estimates of the
regression coefficients, ignoring most of the irrelevant
markers by setting corresponding regression coefficients to zero.
The ridge regression as shown in Figure \ref{fig:sarp}D
does not have the same property of encouraging sparsity
as the lasso. In fact, in statistical literature, it is
well-known that the ridge regression performs poorly in problems that require a selection
of a small number of markers affecting phenotypes.

\begin{table}[t!]
\caption{Summary of results for the association analysis of the asthma dataset}
\label{tbl:sarp}
\begin{center}
\begin{tabular}{c|c|c|c|c|c}
\hline
\multirow{2}{*}{$\rho$} & Number
& \multicolumn{4}{c}{\small Number of nonzero }  \\
& of edges
& \multicolumn{4}{c}{\small regression coefficients}  \\
\cline{3-6}
& & Lasso & \Gc & \Gw & \Gww  \\
\hline
0.3 & 421 & \multirow{4}{*}{125} & 105 &  106 &  108 \\
0.5 & 165 &     & 108 &  107 &  107  \\
0.7 & 71 &  & 105 &  105 &  110  \\
0.9 & 11 &  & 125 &  123 &  123  \\
\hline
\end{tabular}
\end{center}
\end{table}

Since our methods in the GFlasso family
include the lasso penalty, the results from \Gc, \Gw, and \Gww~
show the same property of sparsity as the lasso in their estimates,
as can be seen in Figures \ref{fig:sarp}F-H. In addition,
because of the fusion penalty, the regression
coefficients estimated by our methods form a block structure,
where the regression coefficients for each SNP
are set to the same value within each block.
Thus, each horizontal bar indicates a SNP influencing a correlated
block of phenotypes.  It is clear that the horizontal
bars in Figures \ref{fig:sarp}F-H are generally aligned with
the blocks of highly correlated phenotypes in Figure \ref{fig:sarp}A.
This block structure is much weaker in the results
from the lasso in Figure \ref{fig:sarp}E. For example, Figures
\ref{fig:sarp}F-H show that the SNPs rs3024660 and rs3024622
on the downstream
of Q551R are associated with the same block of traits as Q551R, generating
an interesting new hypothesis that these two SNPs as well as Q551R might be
jointly associated with the same subset of clinical traits.
This block structure shared by the two SNPs is not obvious
in the results of single-marker tests and the lasso.

We fit the lasso and our methods in the GFlasso family,
while varying the threshold for the correlation graph,
and summarize the results in Table \ref{tbl:sarp}.
When the threshold is high at $\rho=0.9$, only a very small number
of edges are included in the phenotype correlation graph,
and the contribution of the graph-guided fusion penalty in GFlasso
is low. Thus, the number of non-zero regression coefficients found
by the \Gc, \Gw, and \Gww~ is similar to the result of the lasso
that does not have the fusion penalty. When we lower the threshold
to $\rho=0.7$, the number of non-zero regression coefficients
decreases significantly for our methods.
As can be seen in Figure \ref{fig:sarp}B,
most of the significant correlation structure is
captured in the thresholded correlation graph at $\rho=0.7$.
Thus, as we further lower the threshold, the number of non-zero
regression coefficients generally remains unchanged.

\section{Discussion}

When multiple phenotypes are involved in association mapping,
it is important to combine the information across phenotypes
and make use of the full information available in data in order to
achieve the maximum power.
Most of the previous approaches either considered each phenotype separately,
or used a two-stage method that first extracts relatively primitive
types of phenotype correlation structure
such as phenotypes transformed through PCA or
subgroups of phenotypes found by clustering algorithms, and then
performs a single-phenotype analysis in the second stage.
Networks or graphs have been extensively studied as a representation
of correlation structure of phenotypes such as gene expression
or clinical traits because they provide a flexible and explicit form
of representation for capturing dependencies. Graphs contain
rich information on phenotype interaction patterns such as
densely connected subgraphs that can be interpreted as a cluster
of phenotypes participating in the same biological process.
Applying a clustering algorithm to identify subgraphs as in
the two-stage methods can potentially result in a loss of information,
decreasing the power of the study. Developing a tool
for multiple-phenotype association mapping that can directly leverage
this full graph structure of phenotypes can offer a way to
combine the large body of previous research in network analysis
with the work on association mapping.

In this article, we proposed a new family of regression methods
called GFlasso that directly incorporates the correlation structure
represented as a graph and uses this information to guide
the estimation process.
The methods consider all of the phenotypes jointly
and estimates the model in a single statistical framework instead of
using a two-stage algorithm. Often, we are interested in detecting
genetic variations that perturb a sub-module of phenotypes rather
than a single phenotype, and the GFlasso achieves this through fusion
penalty in addition to the lasso penalty that encourages parsimony
in the estimated model. The fusion penalty locally
fuses two regression coefficients for a pair of correlated phenotypes,
and this effect propagates through edges of graphs, effectively
applying fusion to all of the nodes within each subgraph.
The ${\rm G_cFlasso}$ used an unweighted graph structure as a guide to
find a subset of relevant covariates that jointly affect highly
correlated outputs. The ${\rm G_wFlasso}$ used additional
information of edge weights to further add flexibility.
Using simulated and asthma datasets, we demonstrated that including
richer information on phenotype structure as in the ${\rm
G_wFlasso}$ and ${\rm G_cFlasso}$ improves the accuracy in detecting
true associations.

We used a simple scheme of a thresholded correlation graph
for learning the correlation structure for phenotypes to be used in the GFlasso.
Many different types of network-learning algortihms have been developed previously.
For example, graphical Gaussian models (GGMs) are constructed based on partial
correlations that capture the direct influence of interacting nodes, and
have been commonly used for inferring gene networks from microarray data.
Furthermore, in order to handle the case with a large number of nodes
and a relatively small sample size, sparse GGMs have been developed.
It would be interesting to see if using more sophisticated
graph learning algorithms can improve the performance of the GFlasso.

In this study, we assumed that the graph structure is available
from pre-processing step. One of the possible extensions of the
proposed method is to learn the graph structure and the regression coefficients
jointly by combining the GFlasso with the graphical lasso~\cite{gplasso}
that learns sparse covariance matrix for phenotypes.
In {\it Geronemo}, both the module network structure and
the markers of regulators
regulating modules were learned simultaneously, although
the genotype information was limited to the markers within regulators~\cite{koller:2006}.
When the GFlasso is extended to learn the graph structure as well
as regression coefficients, the method can be applied to a full
genome-wide scan for associations.

The GFlasso considered only dependencies among phenotypes,
and did not assume any dependencies among the markers.
Since recombination breaks chromosomes during meiosis at non-random sites,
segments of chromosomes are inherited as a unit from ancestors to descendants
rather than an individual nucleotide, creating a relatively low diversity
in observed haplotypes than would be expected if each allele were inherited independently.
Thus, SNPs with high linkage disequilibrium are likely to contribute jointly
to a phenotype in a regression-based penetrance function.
By incorporating the structural information in both the genome and phenome,
we expect to be able to identify a block of correlated markers influencing
a set of correlated phenotypes.

\section*{Appendix: Parameter Estimation}

In this section, we describe the procedure for obtaining estimates
of the regression coefficients in \Gwa.
Since the \Gc~ is a special case of \Gwa~ with $f(r)=1$, the same
procedure can be applied to \Gc~ in a straight-forward manner.
The optimization problem in Equation (\ref{eq:gw_lasso}) is
the Lagrangian form of the following optimization problem:
    \begin{eqnarray}
    {\rm G_wFlasso:} \quad &&
    \hat{\mathbf{B}}^{\textrm{GW}} = \textrm{argmin} \
        \sum_k (\mathbf{y}_k - \mathbf{X}\bm{\beta}_k)^{T}
            \cdot (\mathbf{y}_k - \mathbf{X}\bm{\beta}_k)
        \quad\quad\quad\quad\quad\quad
        \label{eq:gw_lasso_q} \\
        \textrm{s. t.} & &
        \sum_k \sum_j |\beta_{jk}| \le s_1 \,\, \textrm{and} \,\,
        \sum_{(m,l)\in E} f(r_{ml}) \sum_j
        |\beta_{jm}-\textrm{sign}(r_{ml})\beta_{jl}| \le s_2. \nonumber
    \end{eqnarray}
where $s_1$ and $s_2$
are tuning parameters corresponding to $\lambda$ and $\gamma$ in
Equation (\ref{eq:gw_lasso}).

\begin{figure}[t!]\centering
\includegraphics[scale = 0.35, angle=-90]{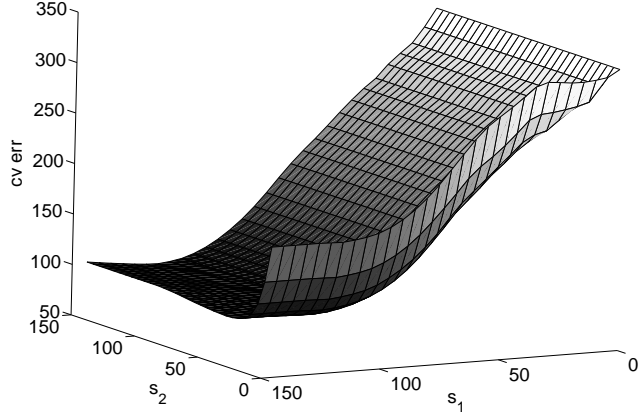}
\caption{
An example of the cross validation error surface 
over a grid of $(s_1,s_2)$ from graph-weighted fused lasso.
}
\label{fig:grid}
\end{figure}

Since the objective function and constraints in Equation (\ref{eq:gw_lasso_q})
are convex, we can formulate this problem as a quadratic programming (QP)
as follows.
Let $\bm{\beta}_{c}$ denote a $(J\!\cdot\! K)$-vector that
can be obtained by concatenating $\bm{\beta}_k$'s such that
$\bm{\beta}_{c} = (\bm{\beta}_1^T, \ldots, \bm{\beta}_K^T)^T$.
We represent ${\beta}_{jk} = {\beta}_{jk}^{+}-{\beta}_{jk}^{-}$, where
${\beta}_{jk}^{+} \ge 0$ and ${\beta}_{jk}^{-} \ge 0$, and let
$\bm{\beta}_c^{+}$ and $\bm{\beta}_c^{-}$ denote $(J\!\cdot\! K)$-vectors of
$\beta_{jk}^{+}$'s and $\beta_{jk}^{-}$'s respectively.
We define $\theta_{j,(m,l)} = \beta_{j,m} - \textrm{sign}(r_{ml})\beta_{j,l}$
for all $(m,l)\in E$ and $j=1,\ldots,J$, and let
$\theta_{j,(m,l)}=\theta_{j,(m,l)}^{+}-\theta_{j,(m,l)}^{-}$
with $\theta_{j,(m,l)}^{+} \ge 0$ and $\theta_{j,(m,l)}^{-} \ge 0$.
Let $\bm{\theta}_c = (\bm{\theta}_{1}^T, \ldots, \bm{\theta}_{|E|}^T)^T$, where
$\bm{\theta}_{e}=(\theta_{1,e}, \ldots, \theta_{J, e})^T$ for $e=(m,l)\in E$.
We define $\bm{\theta}_c^{+}$ and $\bm{\theta}_c^{-}$ similarly.
Let $M$ be a $(J\cdot |E|)\times (J\cdot K)$ matrix, or equivalently
a $|E|\times K$ matrix of $J\times J$ sub-matrices. Each sub-matrix
$\mathbf{B}_{e,k}$ of $M$ for $e=1,\ldots, |E|$ and $k=1,\ldots, K$ is an
identity matrix if $e=(m,l)$ and $k=m$. If $e=(m,l)$ and $k=l$,
$\mathbf{B}_{e,k}$ is set to a diagonal matrix with $-1$ along the diagonal. Otherwise,
$\mathbf{B}_{e,k}$ is set to a matrix of 0's.
Let $R$ be a $(J\cdot |E|)$-vector of $|E|$ sub-vectors with length $J$.
Each sub-vector in $R$ is set to $f(r_{m,l})\cdot \mathbf{1}_J$, where
$\mathbf{1}_J$ represents a $J$-vector of 1's.
Then, the QP problem for Equation (\ref{eq:gw_lasso_q}) can be written as
    \begin{eqnarray}
        \textrm{min}
        \sum_k (\mathbf{y}_k - \mathbf{X}\bm{\beta}_k)^{T}
            \cdot (\mathbf{y}_k - \mathbf{X}\bm{\beta}_k)
    \end{eqnarray}
subject to
    \begin{eqnarray}
    \left(
    \begin{array}{c}
    0 \\
    0 \\
    0 \\
    0 \\
    \end{array} \right)
    \le
    \left(
    \begin{array}{ccccc}
    I & -I & I & 0 & 0 \\
    M & 0 & 0 & -I & I \\
    0 & \mathbf{1}^T_{(J\cdot K)} & \mathbf{1}^T_{(J\cdot K)} & 0 & 0 \\
    0 & 0 & 0 & R^T & R^T \\
    \end{array} \right)
    \left(
    \begin{array}{c}
    \bm{\beta}_c \\
    \bm{\beta}_c^{+} \\
    \bm{\beta}_c^{-} \\
    \bm{\theta}_c^{+} \\
    \bm{\theta}_c^{-}
    \end{array} \right)
    \le
    \left(
    \begin{array}{c}
    0 \\
    0 \\
    s_1 \\
    s_2 \\
    \end{array} \right),
    \end{eqnarray}
where $I$ is a $(J\cdot K)\times (J\cdot K)$ identity matrix, and
$\mathbf{1}_{(J\cdot K)}$ is a $(J\cdot K)$-vector of 1's.

The above QP procedure finds the optimal $\bm{\beta}_k$'s for fixed
$s_1$ and $s_2$. The $s_1$ and $s_2$ can be selected via
cross-validation by running the QP procedure on a grid of $s_1$ and
$s_2$ and selecting the $s_1$ and $s_2$ that give the lowest
validation-set error $C(s_1, s_2)$ as was suggested for the fused
lasso ~\cite{flasso}. The QP solver for ${\rm G_wFLasso}$ runs reasonably fast
for fixed $s_1$ and $s_2$, but the grid search with a
cross-validation can be time-consuming. Instead, we take a
gradient-descent approach that iterates between solving the QP with
the current values of $s_1$ and $s_2$ and updating $s_1$ and $s_2$
with $(s_1,s_2) \leftarrow (s_1,s_2) - \eta \nabla C(s_1,s_2)$,
where the gradient is approximated by a finite difference vector $(
\frac{C(s_1+h, s_2)-C(s_1,s_2)}{h}, \frac{C(s_1,
s_2+h)-C(s_1,s_2)}{h})$.
Figure \ref{fig:grid} shows a typical
example of the cross-validation error over the grid of $(s_1,s_2)$
from ${\rm G_wFlasso}$. We exploit the shape of this error surface,
and determine the initial values $s_1^{(0)}$ and $s_2^{(0)}$ for the
gradient descent as follows. We first search for $s_1^{(0)}$ that
produces the minimum cross-validation error by solving the lasso
with $s_2=\infty$. Then we fix $s_1$ at $s_1^{(0)}$, and perform
another one-dimensional search in the direction of $s_2$, starting from 0
to find the optimal $s_2^{(0)}$ for the ${\rm G_wFlasso}$ along this
path. In our experiments, we found that the initial values obtained
by this procedure was sufficiently close to the global optimum, and
that it converged to the optimum within a relatively small number of
iterations.

\subsubsection*{Acknowledgments}

This material is based upon work supported by an NSF CAREER Award to EPX
under grant No. DBI-0546594, and NSF grants CCF-0523757 and DBI-0640543. EPX
is also supported by an Alfred P. Sloan Research Fellowship of Computer
Science. We thank Sally Winzel, M.D. for providing the SARP dataset, and for
her helpful discussions on the Asthma association study.

\subsection*{Web Resources}

\begin{itemize}
\item
HapMap project: http://www.hapmap.org

\item
Software for GFlasso: http://www.cs.cmu.edu/$\sim$sssykim/gflasso.html
\end{itemize}

\bibliographystyle{ajhg}
\bibliography{assoc2}

\end{document}